%% file: main.tex
\documentclass{article}

\usepackage{arxiv}
\usepackage{times}
\usepackage{epsfig}
\usepackage{graphicx}
\usepackage{amsmath}
\usepackage{amssymb}

\usepackage{microtype}
\usepackage{subfigure}
\usepackage{booktabs} 
\usepackage{amsfonts}
\usepackage{bm}
\usepackage{comment}

\usepackage[numbers,sort,compress]{natbib}

\usepackage{multirow}

\DeclareMathOperator*{\argmax}{\arg\!\max}

\usepackage[pagebackref=true,breaklinks=true,colorlinks,bookmarks=false]{hyperref}

\begin{document}

\title{A Framework using Contrastive Learning for Classification with Noisy Labels}

\author{
  Madalina Ciortan\textsuperscript{\textsection} \\
  EURA NOVA BE\\
  Mont-Saint-Guibert, Belgium\\
  \texttt{madalina.ciortan@euranova.eu} \\
   \And
 Romain Dupuis\textsuperscript{\textsection}\\
  EURA NOVA BE\\
  Mont-Saint-Guibert, Belgium\\
  \texttt{romain.dupuis@euranova.eu} \\
  \And
   Thomas Peel\\
  EURA NOVA BE\\
  Mont-Saint-Guibert, Belgium\\
  \texttt{thomas.peel@euranova.eu} \\
}

\maketitle

\begingroup\renewcommand\thefootnote{\textsection}
\footnotetext{Equal contribution}
\endgroup

\begin{abstract}
We propose a framework using contrastive learning as a pre-training task to perform image classification in the presence of noisy labels.  Recent strategies such as pseudo-labelling, sample selection with Gaussian Mixture models, weighted supervised contrastive learning have been combined into a fine-tuning phase following the pre-training. This paper provides an extensive empirical study showing that a preliminary contrastive learning step brings a significant gain in performance when using different loss functions: non robust, robust, and early-learning regularized. Our experiments performed on standard benchmarks and real-world datasets demonstrate that: i) the contrastive pre-training increases the robustness of any loss function to noisy labels and ii) the additional fine-tuning phase can further improve accuracy, but at the cost of additional complexity.
\end{abstract}


\input{sections/01_intro}

\input{sections/02_related}
\input{sections/03_prelim}
\input{sections/04_representation_classifier}
\input{sections/06_experiments}

\input{sections/07_conclusion}

{\small
\setlength{\bibsep}{3pt}
\bibliographystyle{plainnat}
\bibliography{paper_enx}
}

\pagebreak

\appendix
\section*{Supplementary Materials}

\section{Description of the datasets}
\autoref{tab:description_dataset} gives a detailed description of datasets, including size of the training and test sets, the image resolution, and the number of classes.
\begin{table}[ht]
\begin{center}
\caption{Description of the datasets used in the experiments.}
\label{tab:description_dataset}
\begin{tabular}{p{16mm}p{8mm}p{8mm}p{10mm}p{12mm}}
\hline
Data set       & Train & Test & Size    & \# classes \\ \hline
CIFAR10        & 50K                       & 10K                      & 32x32   & 10         \\ \hline
CIFAR100       & 50K                       & 10K                      & 32x32   & 100        \\ \hline
Clothing1M     & 56K                       & 5K                       & 128x128 & 14         \\ \hline
Mini-Webvision & 66K                       & 2.5K                     & 128x128 & 50         \\ \hline
\end{tabular}
\end{center}
\end{table}

\section{Detailed settings of the experiments}
 All experiments use the ResNet18 as encoder. The classification steps are combined with data augmentation: a random crop with a padding of $4$, an horizontal flip with a probability of $50\%$, and a random rotation of $20^{\circ}$. All other hyperparameters are resumed in~\autoref{tab:hyperparameters}. 
\begin{table}[ht]
\begin{center}
\caption{Training parameters. Symbols: l.r means learning rate, w.d means weight decay, opti. means optimizer, Repre. means representation step, Classi. means the supervised classification step.}
\label{tab:hyperparameters}
\begin{tabular}{p{8mm}p{10mm}p{12mm}p{12mm}p{16mm}}
\cline{3-5}
\multicolumn{1}{l}{}     &        & C10/C100  & Webvision   & Clothing1M \\ \hline
\multirow{5}{*}{Repre.}   & Batch  & 512       & 512         & 512        \\ \cline{2-5} 
                         & Opti.  & Adam      & Adam        & Adam       \\ \cline{2-5} 
                         & l.r.   & $10^{-3}$ & $10^{-3}$   & $10^{-3}$  \\ \cline{2-5} 
                         & w.d.   & $10^{-6}$ & $10^{-6}$   & $10^{-6}$  \\ \cline{2-5} 
                         & epochs & 500       & 500         & 500        \\ \hline
\multirow{5}{*}{Classi.} & Batch  & 256       & 256         & 256        \\ \cline{2-5} 
                         & Opti.  & SGD       & SGD         & SGD        \\ \cline{2-5} 
                         & l.r.   & 0.01/0.1       & 0.4         & 0.01       \\ \cline{2-5} 
                         & w.d.   & $10^{-5}$ & $3.10^{-5}$ & $10^{-4}$  \\ \cline{2-5} 
                         & epochs & 200       & 200         & 200        \\ \hline
\end{tabular}
\end{center}
\end{table}

\section{Ablation study}

\subsection{Contrastive learning with a momentum encoder}
The momentum encoder from the Moco framework~\cite{he2020momentum} maintains a dynamic memory queue of representations. The current mini-batch is added to the memory queue while the oldest mini-batch is dequeued. The offline momentum encoder is a copy of the online encoder by taking an exponentially-weighted average of the parameter of the online encoder.
The main advantage of Moco is to be able to reduce the batch size (and the GPU memory) while keeping a very large number of negative pairs for the contrastive learning.
\begin{table}[ht]
\caption{Top-1 accuracy on CIFAR100 with $80\%$ noise. Two different contrastive learning frameworks are evaluated for the pre-training: SimCLR and Moco. The third column gives the accuracy for a classifier with a smaller learning rate.}
\begin{center}
\begin{tabular}{cccc}
\cline{2-4}
        & SimCLR & Moco & \begin{tabular}[c]{@{}l@{}}Moco - Fine\\ tune\end{tabular} \\ \hline
CE      & 12.4   & 12.0 & 49.0                                                       \\ \hline
ELR     & 45.3   & 38.8 & 42.3                                                       \\ \hline
NFL+RCE & 50.2   & 26.3 & 47.0                                                         \\ \hline
\end{tabular}
\end{center}
\label{tab:moco_simclr}
\end{table}

The different representations computed by SimCLR and Moco are compared on CIFAR100. Both approaches are trained for 500 epochs following the usual hyperparameter parameters from the initial papers. As the two methods use different strategies to compute the representations, their quality is assessed by learning a linear classifier on top of the frozen encoder network. It can be seen as a proxy for representation quality. The SimCLR framework reaches $55.3\%$ of accuracy while Moco gets $55.0\%$ of accuracy. However, the two encoders do not behave in a similar way with regard to noisy labels. The same classifier (multi-layer, same learning rate and weight decay) is trained starting from the representation computed by SimCLR and Moco. As depicted in~\autoref{tab:moco_simclr}, the representations computed by Moco are more sensitive to the noisy labels. However, reducing the learning rate of the optimizer by a factor $10$ (column Moco - Fine Tune) significantly increases the accuracy. 

Even if pretraining the encoder increases the accuracy for both contrastive methods, the two approaches do not have the same behavior. In particular, the best parameters for the classifier optimizer seem to be different. This raises several questions about the difference between the two representations and what properties of these representations improve the robustness of the classifier.


\subsection{Sensitivity to the learning rate}
\begin{figure}[]
\vskip 0.2in
\begin{center}
\subfigure[CIFAR100 with $80\%$ noise.]{\includegraphics[width=0.5\columnwidth]{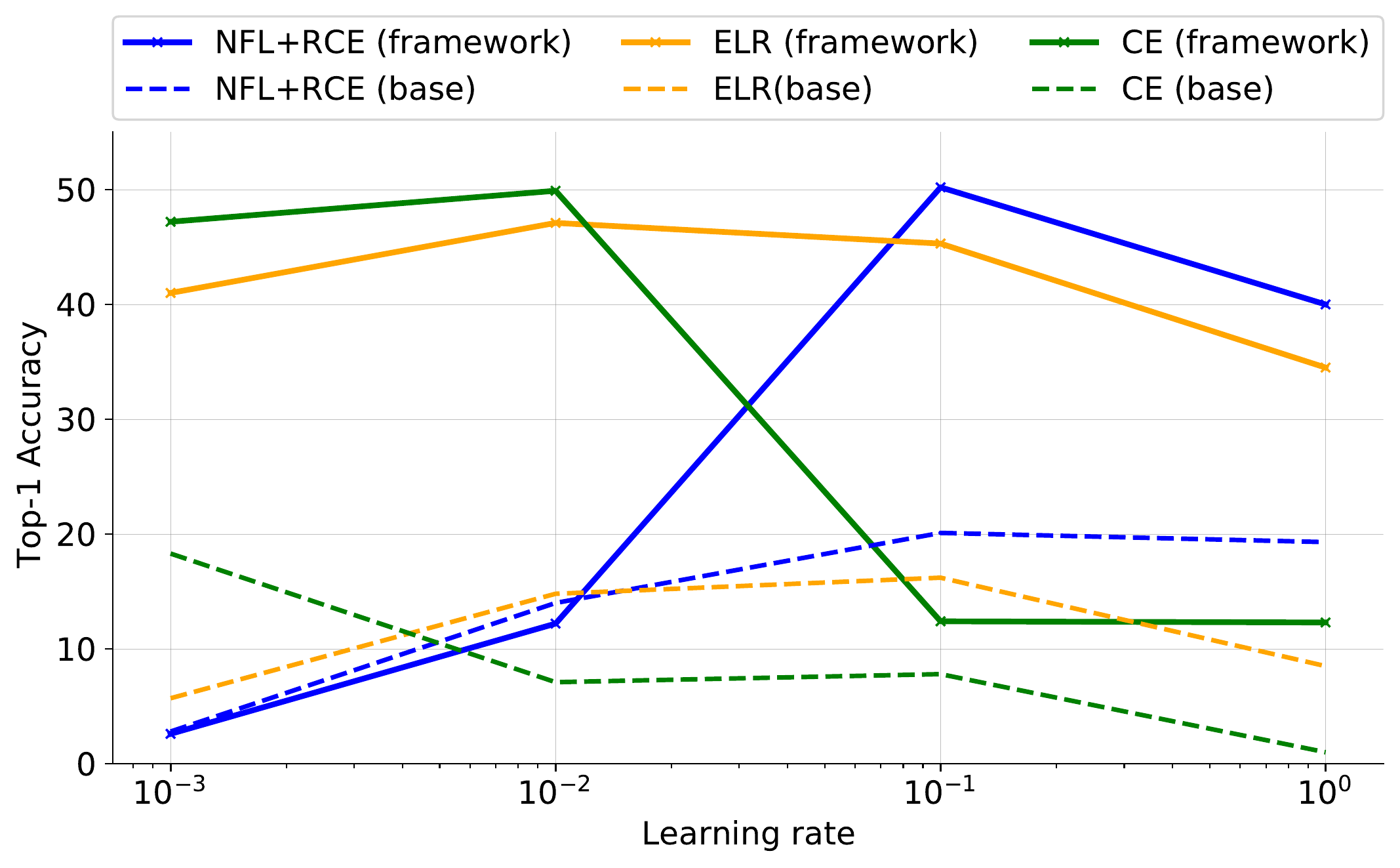}}
\subfigure[CIFAR100 with $60\%$ noise.]{\includegraphics[width=0.5\columnwidth]{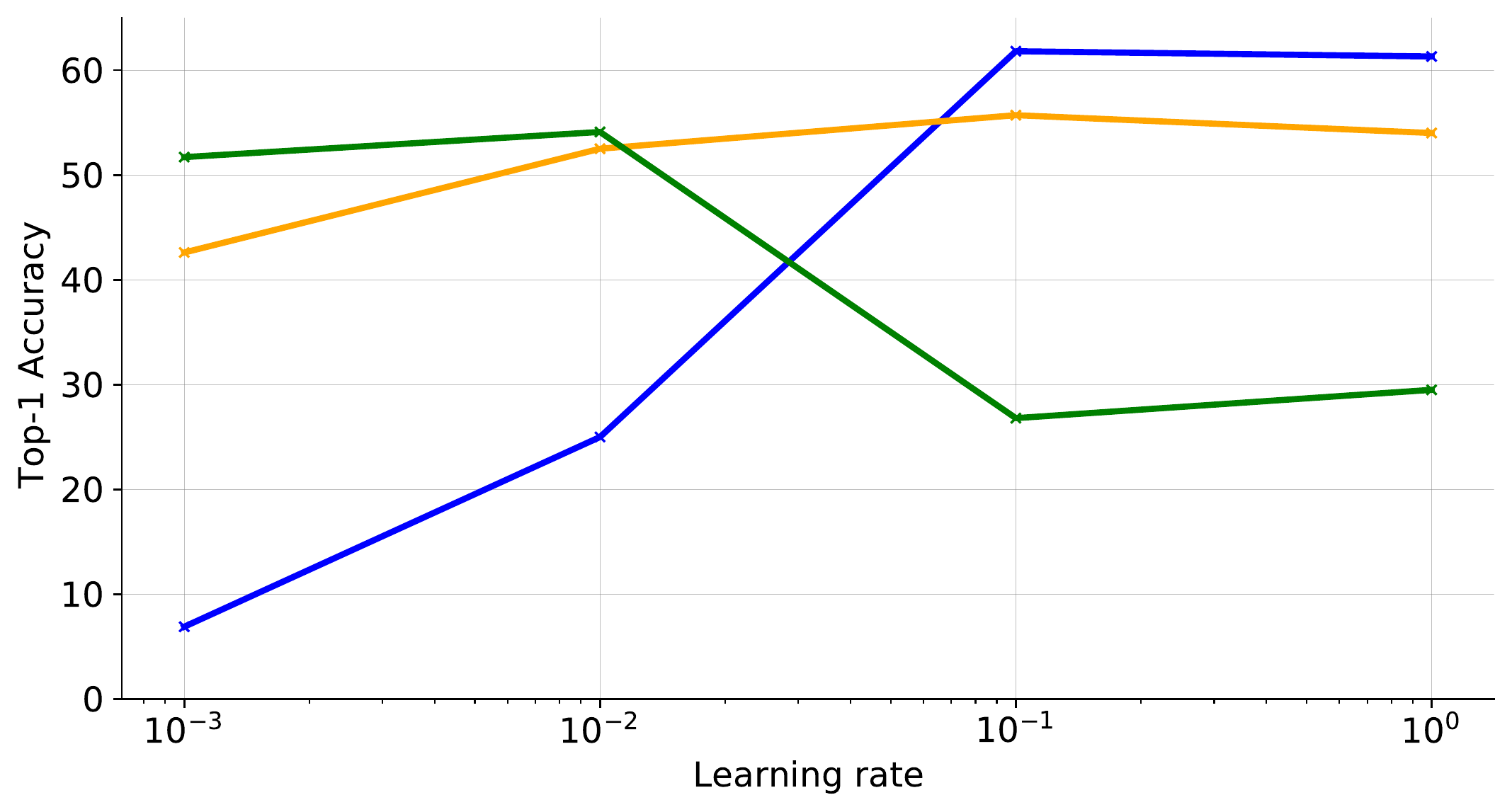}}
\subfigure[CIFAR100 with $40\%$ noise.]{\includegraphics[width=0.5\columnwidth]{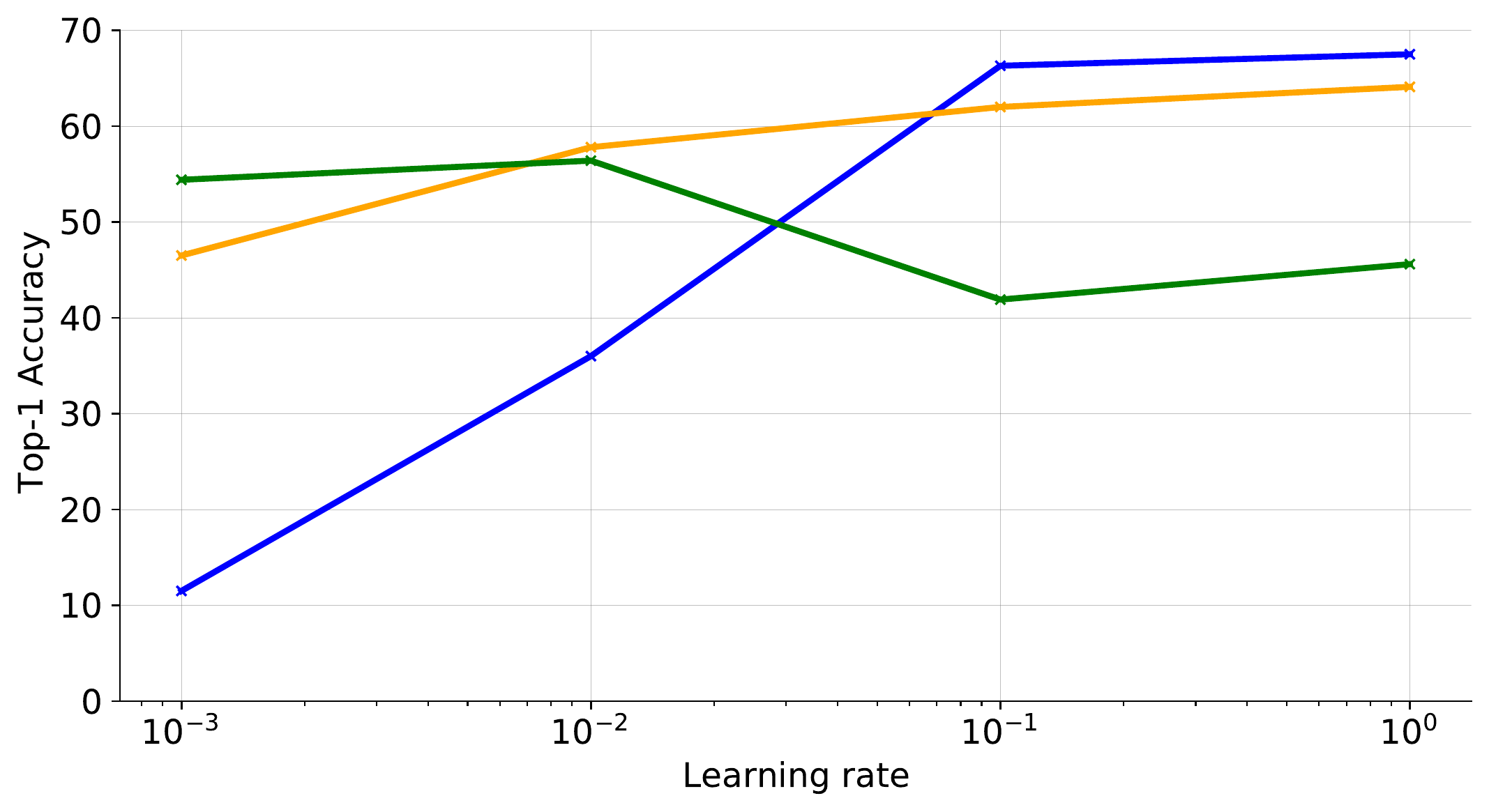}}
\subfigure[CIFAR100 with $20\%$ noise.]{\includegraphics[width=0.5\columnwidth]{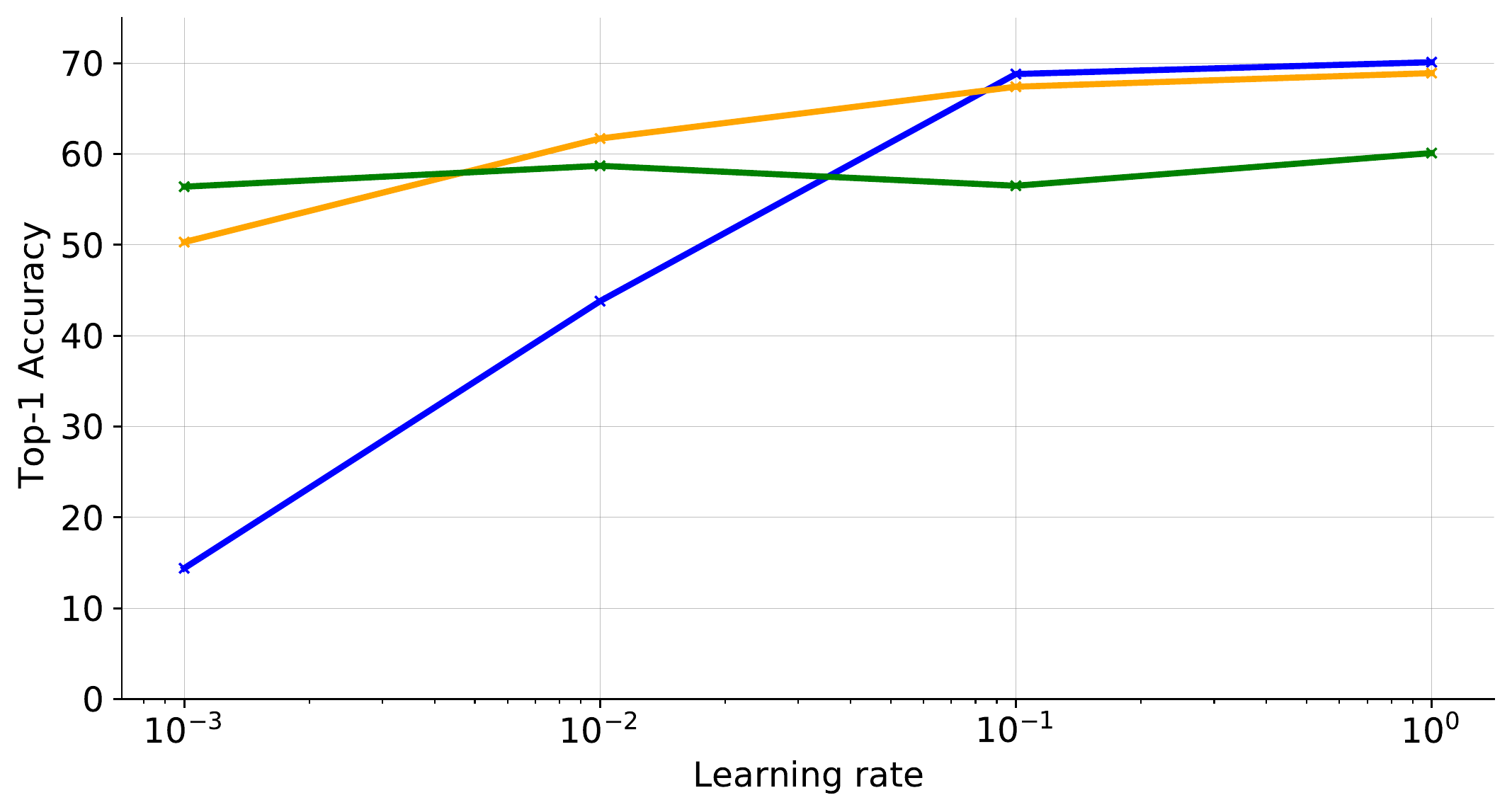}}
\caption{Hyperparameter sensitivity for CIFAR100.}
\label{fig:sensitivity2}
\end{center}
\vskip -0.2in
\end{figure}

We perform an hyperparameter search on the CIFAR100 datasets. The learning rate is chosen in $\{10^{-3}, 10^{-2}, 10^{-1}, 10^{0} \}$. Results are presented in~\autoref{fig:sensitivity2}.
The configuration with $80\%$ noise is clearly the most sensitive case, in particular for the NFL+RCE loss and the CE. The ELR method is quiet robust over the investigated range. 

\subsection{Impact of the classifier architecture}
The impact of the 2 classifier architectures is detailed in ~\autoref{tab:results_classifier_arch}. The multilayer architecture performs better on datasets contaminated with a significant amount of asymmetric noise.
\begin{table}[ht]
\caption{Results on both CIFAR10 and CIFAR100 using symmetric noise (0.2 - 0.8) and asymmetric noise (0.2 - 0.4). We compare a single linear layer (L) to multiple layers (M) final classification head, for three losses: CE, ELR, and NFL+RCE.}
\label{tab:results_classifier_arch}
\vskip 0.15in
\begin{center}
\begin{tabular}{p{6.0mm}p{4mm}p{6.5mm}llll}
\cline{4-7}
                      &                         &          & \multicolumn{2}{c}{CIFAR10} & \multicolumn{2}{c}{CIFAR100}  \\ \hline
Type                  & $\eta$ & Loss     & L    & M         & L         & M      \\ \hline
\multirow{12}{*}{Sym} & \multirow{3}{*}{0.2}    & ce       & \textbf{91.7}     & 87.7    & \textbf{58.6 }         & 56.5 \\ \cline{3-7} 
                      &                         & elr      & 92.9     & \textbf{93.0}    & 66.4         & \textbf{67.4} \\ \cline{3-7} 
                      &                         & nfl\_rce & \textbf{93.2 }    & 92.7    &\textbf{ 69.7 }         & 68.8 \\ \cline{2-7} 
                      & \multirow{3}{*}{0.4}    & ce       & \textbf{90.6 }   & 78.0    & \textbf{ 44.2    }      & 41.9 \\ \cline{3-7} 
                      &                         & elr      & \textbf{92.1}    & 92.0    & 60.8          & \textbf{62.0} \\ \cline{3-7} 
                      &                         & nfl\_rce & \textbf{92.1 }    & 91.4    & \textbf{ 67.0 }          & 66.3 \\ \cline{2-7} 
                      & \multirow{3}{*}{0.6}    & ce       & \textbf{88.1}    & 59.2    & \textbf{28.9}          & 26.8 \\ \cline{3-7} 
                      &                         & elr      & 89.7     & \textbf{90.4}    & 54.0          & \textbf{55.7} \\ \cline{3-7} 
                      &                         & nfl\_rce &\textbf{ 90.2  }   & 88.1    & \textbf{63.7   }       & 61.8 \\ \cline{2-7} 
                      & \multirow{3}{*}{0.8}    & ce       & \textbf{72.6}     & 27.3    & \textbf{14.1}          & \textbf{12.4} \\ \cline{3-7} 
                      &                         & elr      & 82.0     & \textbf{84.8}    & 41.6         & \textbf{45.3} \\ \cline{3-7} 
                      &                         & nfl\_rce &\textbf{ 78.9 }    & 59.9    &\textbf{ 54.2 }         & 50.2 \\ \hline
\multirow{9}{*}{Asym} & \multirow{3}{*}{0.2}    & ce       & \textbf{91.6 }    & 87.9    & \textbf{60.1} & 57.8          \\ \cline{3-7} 
                      &                         & elr      & \textbf{92.7 }    & 92.4    & 69.3 & \textbf{70.2 }         \\ \cline{3-7} 
                      &                         & nfl\_rce &\textbf{ 92.5}     & 91.5    & \textbf{69.1 }         & 68.4 \\ \cline{2-7} 
                      & \multirow{3}{*}{0.3}    & ce       & \textbf{90.2}     & 83.9    & \textbf{52.3} & 50.4          \\ \cline{3-7} 
                      &                         & elr      & 90.6     & \textbf{91.7}    & 68.5 & \textbf{69.3 }         \\ \cline{3-7} 
                      &                         & nfl\_rce & \textbf{91.2}     & 89.9    & \textbf{68.0   }       & \textbf{63.5} \\ \cline{2-7} 
                      & \multirow{3}{*}{0.4}    & ce       & \textbf{84.7  }   & 77.8    & \textbf{43.7}          & 42.4 \\ \cline{3-7} 
                      &                         & elr      & 68.4     & \textbf{89.5}    & 65.5 & \textbf{67.6} \\ \cline{3-7} 
                      &                         & nfl\_rce & 62.6     & \textbf{82.4}    & \textbf{63.0 }         & 47.8 \\ \hline
\end{tabular}
\end{center}
\vskip -0.15in
\end{table}

\section{Dynamic bootstrapping with mixup}
In addition to the presented fine-tuning phase, we also evaluated the performance of other techniques recently proposed for noisy label classification. The weights $w$ computed by the sample selection phase can also be combined with a mixup data augmentation strategy~\cite{Zhang2018mixup}. A specific strategy for noisy labels, called dynamic bootstrapping with mixup~\cite{arazo2019unsupervised}, has been developed to help convergence under extreme label noise conditions. The convex combinations of sample pairs $\bm{x_p}$ (loss $l_p$) and $\bm{x_q}$ (loss $l_q$) is weighted by the probability $w$ to belong to the clean dataset:
\begin{equation}
    \bm{x} = \frac{w_p} {w_p + w_q}\bm{x_p} + \frac{ w_q}{ w_p + w_q }\bm{x_q}.
\end{equation}
\begin{equation}
    l = \frac{w_p} {w_p + w_q}l_p + \frac{w_q}{ w_p + w_q }l_q.
\end{equation}
The associated CE is corrected according to the weights:
\begin{equation}
    l_{ce} = - \sum_{k=1}^{K} \left(w_i q(k | \bm{x_i}) + (1-w_i) z_i\right) log(p(k | \bm{x_i})),
\end{equation}
where $z(k | \bm{x_i}) = 1$ if $k=\argmax p(k | \bm{x_i})$ or zero for all the other cases. If the GMM probability are well estimated, combining one noisy sample with one clean sample leads to a large weight for the clean sample and a small weight for the noisy sample. Clean-clean and noisy-noisy cases remain similar to a classical mixup with weights around $0.5$.

The dynamic bootstrapping for ELR is derived by replacing the CE term by the corrected version:
\begin{equation}
l_{elr}(\theta) = l_{ceb}(\theta) + \frac{\lambda_{elr}}{N} log \left( 1 - \sum_{k=1}^{K} p(k|\bm{x_i}). t(k|\bm{x_i} )     \right). 
\end{equation}

Regarding the robust loss function NFL+RCE, the two losses have to be modified:
\begin{equation}
\begin{aligned}
    & l_{nfl} =  w_i \frac{ -\sum\limits_{k=1}^{K} q(k|\bm{x_i}) (1-p(k|\bm{x_i}))^\gamma log(p(k|\bm{x_i}))}{-\sum\limits_{j=1}^K \sum\limits_{k=1}^{K} q(y=j|\bm{x_i}) (1-p(k|\bm{x_i}))^\gamma log(p(k|\bm{x_i}))} \\
    & + (1 - w_i) \frac{ -\sum\limits_{k=1}^{K} z(k|\bm{x_i}) (1-p(k|\bm{x_i}))^\gamma log(p(k|\bm{x_i}))}{-\sum\limits_{j=1}^K \sum\limits_{k=1}^{K} z(y=j|\bm{x_i}) (1-p(k|\bm{x_i}))^\gamma log(p(k|\bm{x_i}))}
\end{aligned}
\end{equation}
where $q$ is the one-hot encoding of the label (the zero value is fixed to a low value to avoid $log(0)$).
\begin{equation}
    l_{rce} = -\sum_{k=1}^{K} p(k |\bm{x_i})log \left( w_i. q(k | \bm{x_i}) + (1-w_i)z_i \right)
\end{equation}

\section{Classification warmup}
\begin{figure}[ht]
\vskip 0.2in
\begin{center}
\centerline{\includegraphics[width=\columnwidth]{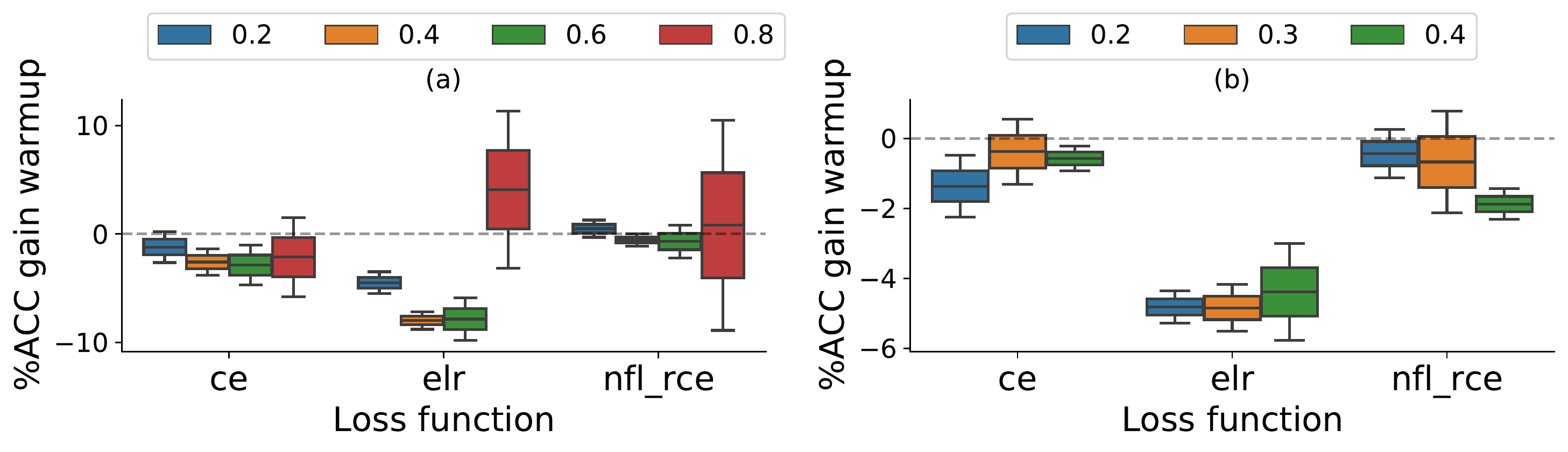}}
\caption{Gain in performance when using a supplementary classifier warm-up phase before training the entire model on CIFAR 100 with symmetric (panel a) and asymmetric noise (panel b).  } 
\label{fig:warmup}
\end{center}
\vskip -0.2in
\end{figure}

This section compares the classification accuracy of models trained with and without a warm-up phase after the representation learning. The warm-up phase consists of freezing the entire model except for the classification head.~\autoref{fig:warmup} depicts the gain in performance brought by the warm-up phase. When using the default values, its inclusion is beneficial only for significant amounts of symmetric noise. Our experiments have been performed using only the recommended classifier learning rates, detailed in the experimental setup. Having different learning rates for the warm-up phase and the classification optimizing all weights (encoder and classifier) could have a different impact on the warmup phase. 

\section{Execution time analysis}

In order to estimate our method's computational cost, we compared the execution time of both approaches, consisting of performing only the pre-training phase and the pre-training followed by fine-tuning with the execution time of performing only one supervised classification phase (i.e. the baseline). The number of times our methods were slower than the baseline has been depicted in ~\autoref{tab:execution_times}. We provided similar metrics for the methods making available this informations (i.e. Taks, Co-teaching+, JoCoR). As expected, the pre-training doubles the execution time of the baseline as, in addition to training the classifier, a contrastive learning phase has to be performed beforehand. The entire framework introduces a computational cost 3 to 4.5 times higher. However, all methods leveraging pre-trained models (using for instance supervised pre-training) also hide a similar computational cost.

\begin{table}[]
\caption{Comparison of execution time results reported as a factor with respect to the training time of the baseline, representing the supervised training of the model with the CE loss. The abbreviations Ours (Pre-t) indicate the pre-training phase while Ours (Fine-tune) the pre-training phase followed by fine-tuning}
\label{tab:execution_times}
\begin{center}
\begin{tabular}{lllll}
\hline
                                                        & \begin{tabular}[c]{@{}l@{}}C10\\ 80\% S\end{tabular} & \begin{tabular}[c]{@{}l@{}}C10\\ 40\% A\end{tabular} & \begin{tabular}[c]{@{}l@{}}C100\\ 80\% S\end{tabular} & \begin{tabular}[c]{@{}l@{}}C100\\ 40\% A\end{tabular} \\ \hline
\begin{tabular}[c]{@{}l@{}}Ours \\ (Pre-t)\end{tabular} & 2.36                                                 & 2.53                                                 & 2.40                                                  & 2.32                                                  \\ \hline
\begin{tabular}[c]{@{}l@{}}Ours \\ (Fine-tune)\end{tabular}  & 3.42                                                 & 3.63                                                 & 4.31                                                  & 4.36                                                  \\ \hline
Taks                                                    & 0.53                                                 & 1.04                                                 & 0.52                                                  & 0.98                                                  \\ \hline
Co-teach+                                               & 2.00                                                 & 2.00                                                 & 2.00                                                  & 2.01                                                  \\ \hline
JoCoR                                                   & 1.73                                                 & 1.74                                                 & 1.72                                                  & 1.74                                                 
\end{tabular}
\end{center}
\end{table}

\section{An attempt to prevent overfitting with early stopping}
Overfitting is the common weakness of all studied models. Several strategies understanding and preventing overfitting have been explored: i) analysing the model behaviour on a validation set, 
ii) identifying the start of the memorization phase using Training  Stop Point~\cite{kamabattula2020identifying}, and 
iii) characterizing changes in the model using Centered Kernel Alignment~\cite{kornblith2019similarity}. A clean validation set is generally used to find the best moment for early stopping and to estimate the hyperparameter sets. However, we assume that clean validation samples are not available. Therefore, the methods must be robust to overfitting and to a wide range of hyperparameter values.

\begin{figure}[ht]
\vskip 0.2in
\begin{center}
\centerline{\includegraphics[width=\columnwidth]{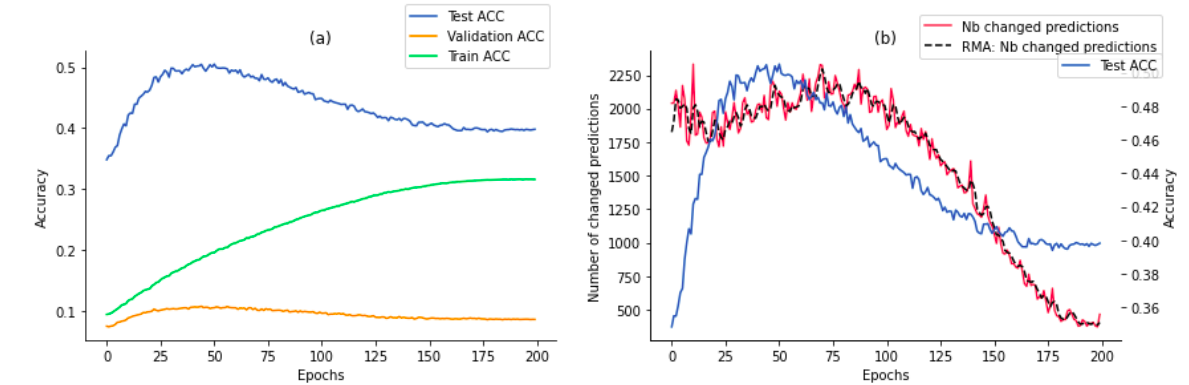}}
\caption{ Evolution of accuracy across train/validation/test sets (a) Prediction stability on the validation set computed as the number of samples changing class across consecutive epochs. The rolling mean average of the number of predictions has been depicted in black. The experiments have been performed on CIFAR 100, with 80\% symmetric noise during the first classification phase and used NFL + RCE loss.}
\label{fig:validation}
\end{center}
\vskip -0.2in
\end{figure}

As typical noisy label settings lack a clean reference set, we contrasted the behavior of the model on a corrupted validation set with that on a clean test set, where overfitting can be easily identified. Train/validation sets have been generated using 5 cross validation folds. In the figure below, panel (a) depicts the evolution of accuracy scores on the corrupted train/validation sets as well as on the test set. After the first 50 epochs, the model starts overfitting as the test accuracy drops by 10\% (~\autoref{fig:validation} panel a). The accuracy on the corrupted train continues to increase as the model memorizes the input labels. However, on the corrupted validation set a plateau followed by a loss of performance is indicative of the same phenomena, but without being always aligned with the overfitting phase observed on the test-set. The memorization phenomena of the train-set labels incapacitates the model to generalize on the corrupted validation set and explains the significant difference in scores between the train and validation accuracies.

A second perspective on the analysis of overfitting explores the stability of the network's predictions on the validation set. Panel (b) depicts the number of samples predicted in different classes across consecutive epochs. As the model starts overfitting, the prediction stability also increases. After 200 epochs, only 500 from 10000 samples on the validation set change class from one epoch to another. As expected, the network stability is correlated with model overfitting on severe label noise.

\begin{figure}[ht]
\vskip 0.2in
\begin{center}
\centerline{\includegraphics[width=\columnwidth]{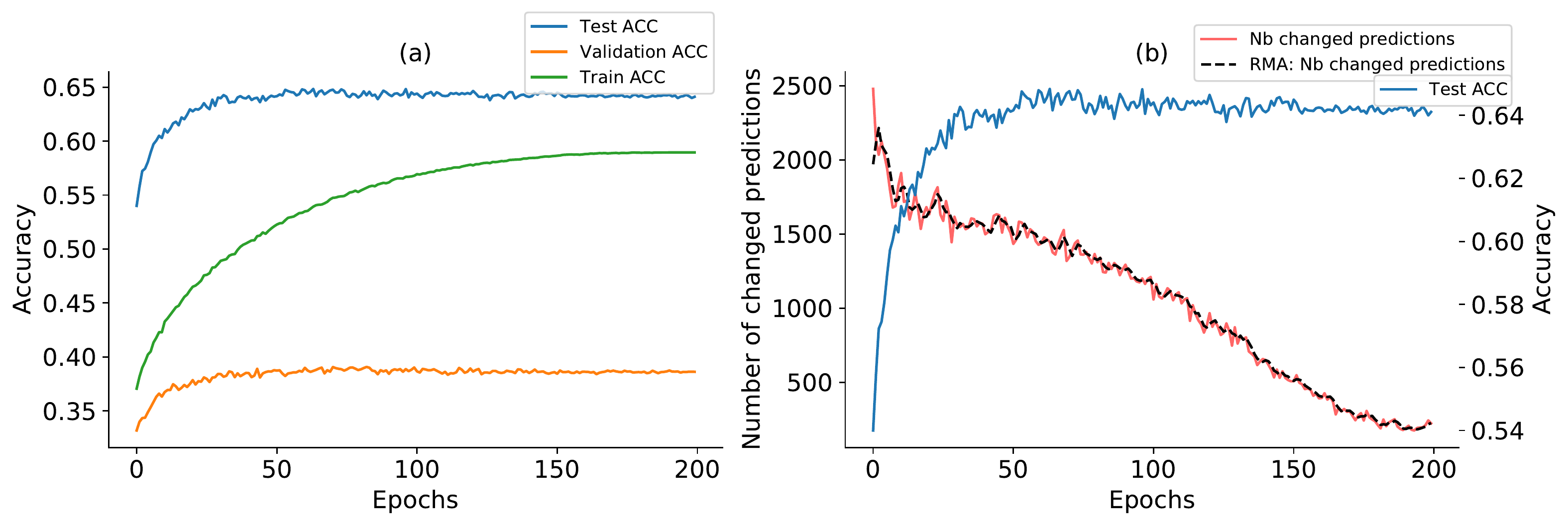}}
\caption{ Evolution of accuracy across train/validation/test sets (a) Prediction stability on the validation set computed as the number of samples changing class across consecutive epochs. The rolling mean average of the number of predictions has been depicted in black. The experiments have been performed on CIFAR 100, with 40\% asymmetric noise during the first classification phase and used NFL + RCE loss.}
\label{fig:validation_asym}
\end{center}
\vskip -0.2in
\end{figure}

\begin{figure}[ht]
\vskip 0.2in
\begin{center}
\centerline{\includegraphics[width=0.7\columnwidth]{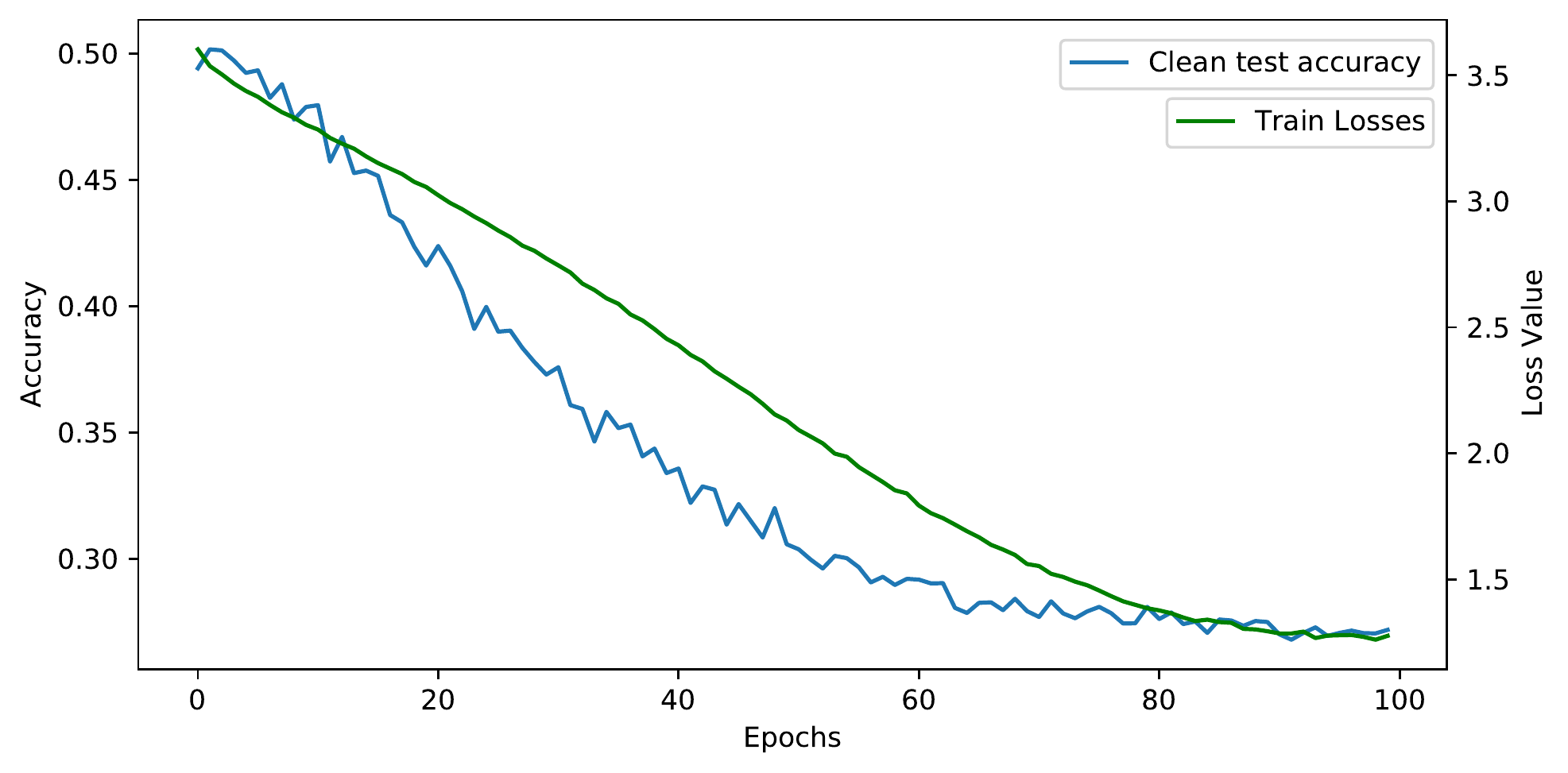}}
\caption{Evolution of train loss and test accuracy on CIFAR, 60\% symmetric noise. The theoretical conditions of higher variance on the train loss, associated with the start of the memorization phase, as suggested by TSP, are not fulfilled.  } 
\label{fig:TPS}
\end{center}
\vskip -0.2in
\end{figure}

Several recent contributions studied the overfitting phenomena of neural networks in an attempt to identify an early stopping point corresponding to the maximum obtainable test accuracy. Traditional approaches leverage a clean test set which is often unavailable when confronted with noisy labelled data.~\citet{kamabattula2020identifying} proposed to find a Training Stop Point (TSP), a heuristic analyzing the rate of change in the training accuracy and correlated its transition towards the memorization phase with a transition towards a smoother (smaller variance) regime, as depicted below. Our experimental results showed that the theoretical conditions to identify the early stopping point are not always met as suggested by TSP. ~\autoref{fig:TPS} indicates that the overfitting phase, starting after the first 5 epochs, does not change the variance of the train loss.

Centered Kernel Alignment (CKA)~\cite{kornblith2019similarity} provides a similarity index comparing representations between layers of different trained models. In particular, CKA shows interesting properties as CKA can consistently identify correspondences between layers trained from different initializations.

\begin{figure}[ht]
\vskip 0.2in
\begin{center}
\subfigure[CKA from a pre-trained encoder with contrastive learning.]{\includegraphics[width=0.7\columnwidth]{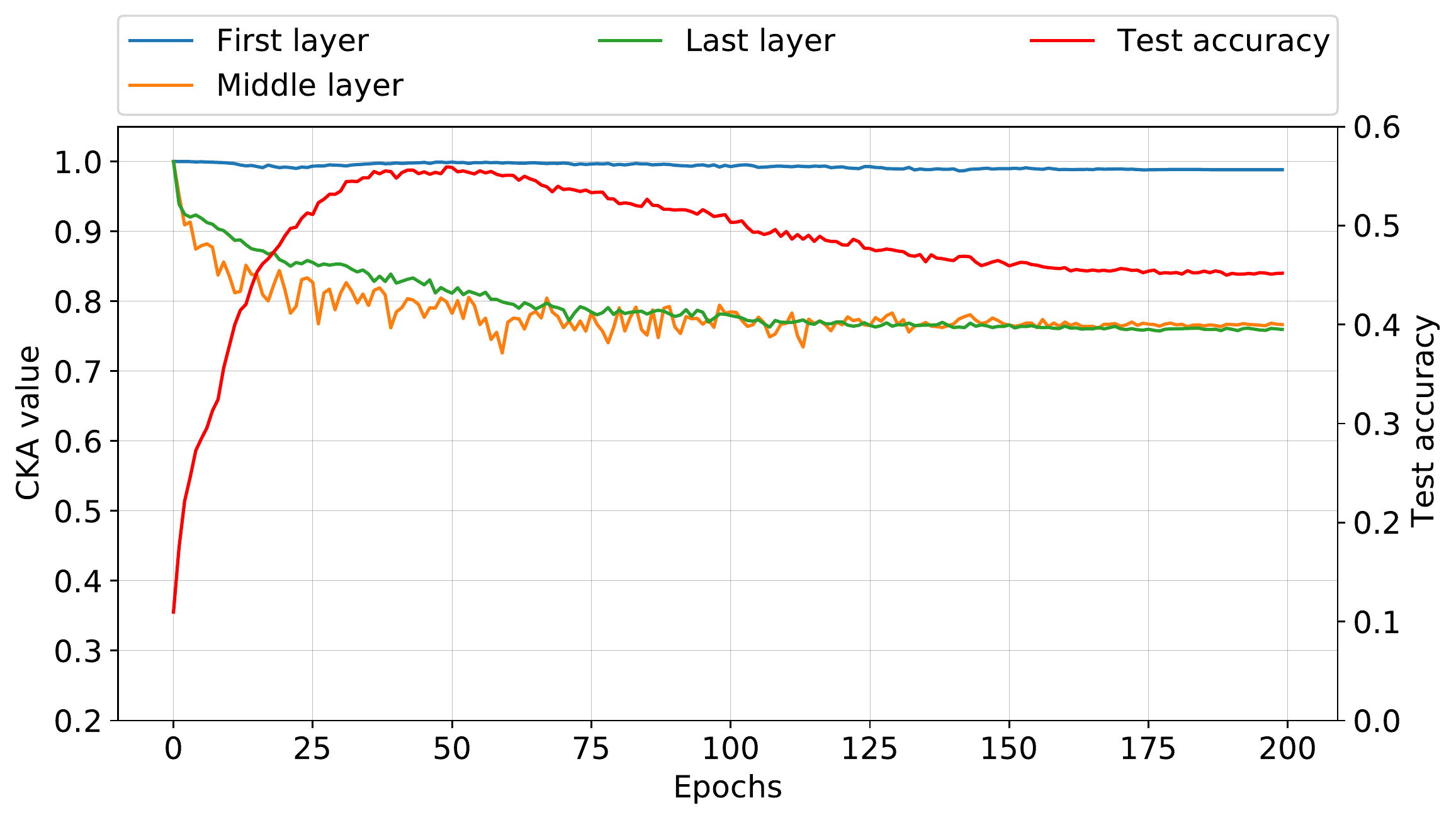}}
\subfigure[CKA from a random initialization.]{\includegraphics[width=0.7\columnwidth]{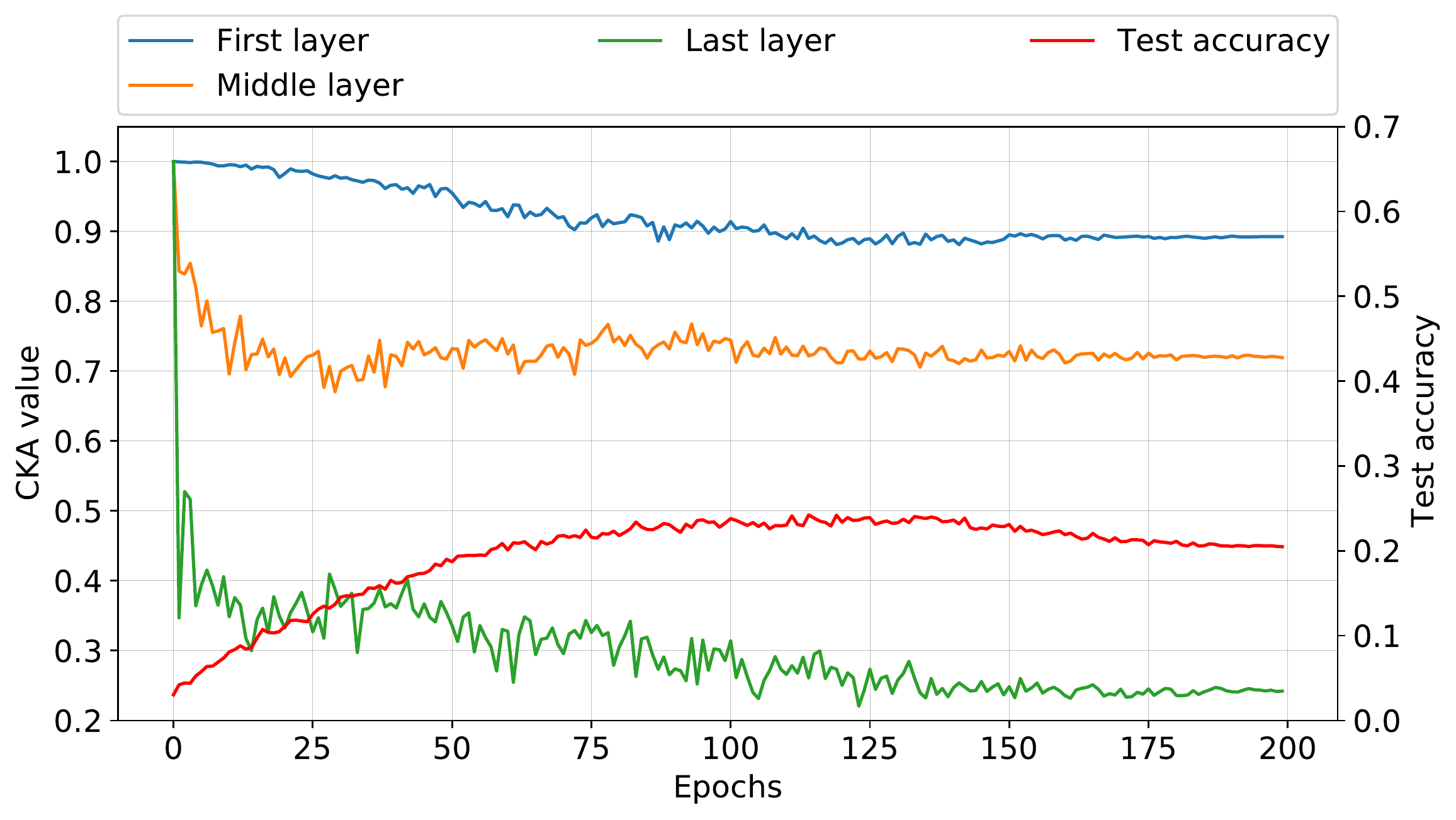}}
\caption{CKA similarity for a model trained with NFL+RCE loss function on CIFAR100 with $80\%$ noise.}
\label{fig:cka}
\end{center}
\vskip -0.2in
\end{figure}

The objective is twofold: i) observing if a specific behavior can be identified for the overfitting and ii) comparing the CKA values with and without contrastive pre-training.  The CKA index is computed at three different locations in the network: the input layer, the middle of the network, and the final layer.~\autoref{fig:cka} shows the CKA similarity computed between the initialization/pre-trained model and the same layer at different epochs during the training process.
It is interesting to note that the first layer of the pre-trained model remains very similar to the same layer computed by contrastive learning. Such behavior was expected in order to improve the robustness against noisy labels. Indeed, if contrastive learning can extract good representations for semi-supervised or transfer learning, being close to such representations can also help to avoid learning noisy labels. As expected, all layers of the model trained from a random initialization vary much more during the training.

The training phase of the pre-trained model reaches its maximum accuracy around 50 epochs but the CKA values of the middle and last layers continue to drop until 130 epochs. On the other hand, the CKA values of the initialized model remain stable after $150$ epochs when the test accuracy reaches almost its maximum value. At first glance, the CKA behavior cannot be related to overfitting. 

None of the studied approaches provides a solution preventing overfitting across all our experiments and this problem remains an open question.

\end{document}

%% file: sections/01_intro.tex
\section{Introduction}
Collecting large and well-annotated datasets for image classification tasks represents a challenge as human quality annotations are expensive and time-consuming. Alternative methods exist, such as web crawlers~\cite{mahajan2018exploring}. Nevertheless, these methods generate noisy labels decreasing the performance of deep neural networks. They tend to overfit to noisy labels due to their high capacity~\cite{zhang2016understanding}. That is why developing efficient noisy-label learning (NLL) techniques is of great importance.

\begin{figure}[ht]
\vskip 0.1in
\begin{center}
\centerline{\includegraphics[width=0.7\columnwidth]{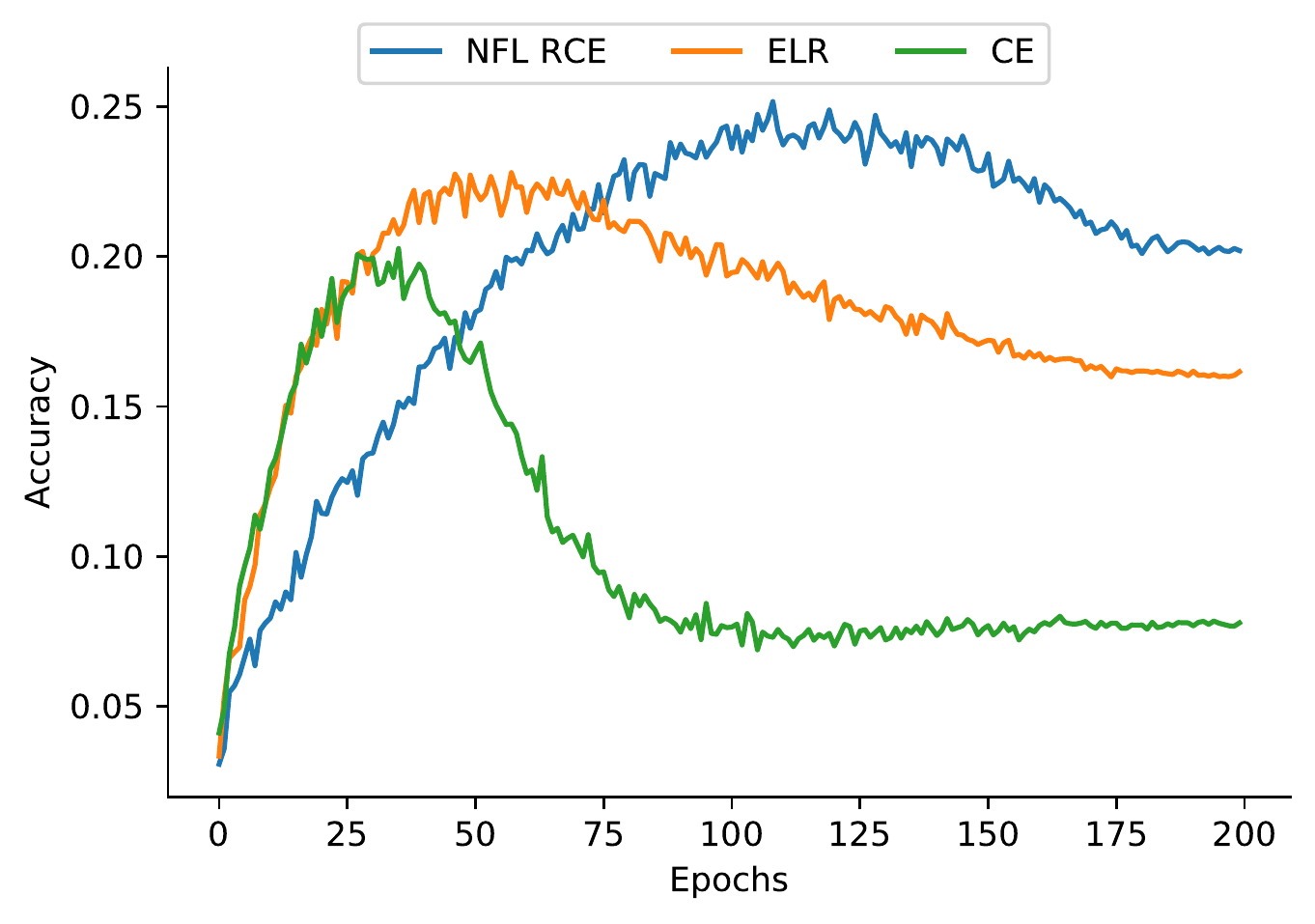}}
\caption{Top-1 test accuracy for a ResNet18 trained on the CIFAR-100 dataset with a symmetric noise of 80\% for three losses: Cross Entropy (CE), Normalized Focal Loss + Reverse Cross Entropy (NFL+RCE), and Early Learning Regularization (ELR).}
\label{fig:broad_loss_comparison}
\end{center}
\vskip -0.2in
\end{figure}

Various strategies have been proposed to deal with NLL: i) Noise transition matrix~\cite{patrini2017making, goldberger2017TrainingDN, xia2019anchor} estimates the noise probability and corrects the loss function, ii) a small and clean subset can help to avoid overfitting~\cite{hendrycks2018using}, iii) samples selection identifies true-labeled samples~\cite{jiang2018mentornet, han2018co,li2020dividemix}, and iv) robust loss functions solve the classification problem only by adapting the loss function to be less sensitive to noisy labels~\cite{zhang2018generalized, wang2019symmetric, ma2020normalized}. Methods also combine other strategies (eg. ELR+~\cite{li2020dividemix}, DivideMix~\cite{liu2020early}): two networks, semi-supervised learning, label correction, or mixup. They show the most promising results but lead to a large number of hyperparameters. That is why we explore improvement strategies for robust loss functions. They are simpler to integrate and faster to train, but as illustrated in~\autoref{fig:broad_loss_comparison}, they tend to overfit and have lower performance for high noise ratios.

Meanwhile, new self-supervised learning algorithms for image representations have been recently developed~\cite{chen2020simple, he2020momentum}. Such algorithms extract representation (or features) in unsupervised settings. These representations can then be used for downstream tasks such as classification. Methods based on contrastive learning compete with fully supervised learning while fine-tuning only on a small fraction of all available labels. Therefore, using contrastive learning for NLL appears as promising. In this work, contrastive learning aims to pre-train the classifier to improve its robustness.

The key contributions of this work are:
\begin{itemize}
    \item A framework increasing robustness of any loss function to noisy labels by adding a contrastive pre-training task.
    
    \item The adaptation of the supervised contrastive loss to use sample weight values, representing the probability of correctness for each sample in the training set
    
    \item An extensive empirical study identifying and benchmarking additional state of the art strategies to boost the performance of pre-trained models: pseudo-labeling, sample selection with GMM, weighted supervised contrastive learning, and mixup with bootstrapping.
\end{itemize}

%% file: sections/02_related.tex
\section{Related works}
\label{sec:related_works}
Existing approaches dealing with NLL and contrastive learning in computer vision are briefly reviewed. Extra details can be found in ~\citet{song2020learning, le2020contrastive}.

\subsection{Noise tolerant classification}
\textbf{Sample Selection}: This method identifies noisy and clean samples within the training data. Several strategies leverage the interactions between multiple networks to identify the probably correct labels~\cite{han2018co, jiang2018mentornet, li2020dividemix}. Recent works~\cite{arazo2019unsupervised, song2019selfie} exploit the small loss trick to identify clean and noisy samples by considering a certain number of small-loss training samples as true-labeled samples. This approach can be justified by the memorization effect: deep neural networks first fit the training data with clean labels during a so-called early learning phase, before overfitting the noisy samples during the memorization phase~\cite{arpit2017closer, liu2020early}. 

\textbf{Robust Loss Function}: Commonly used loss functions, such as Cross Entropy (CE) or Focal Loss, are not robust to noisy labels. Therefore, new loss functions have been designed. Such robust loss functions can be easily incorporated into existing pipelines to improve performance regarding noisy labels. The symmetric cross entropy~\cite{wang2019symmetric} has been proposed by adding a reverse CE loss to the initial CE. This combination improves the accuracy of the model compared to classical loss functions.~\citet{ma2020normalized} show theoretically that normalization can convert classical loss functions into loss functions robust to noise labels. The combination of two robust loss functions can also improve robustness. However, the performance of normalized loss functions remains quite low for high noise rates as illustrated in~\autoref{fig:broad_loss_comparison}.

\textbf{Semi-supervised}: Semi-supervised approaches deal with both labeled and unlabeled data. Recent works~\cite{nguyen2019self, li2020dividemix, wang2020seminll} combine sample selection with semi-supervised methods: the possibly noisy samples are treated as unlabeled and the possibly clean samples are treated as labeled. Such approaches leverage information contained in noisy data, for instance by using MixMatch~\cite{berthelot2019mixmatch}. Semi-supervised approaches show competitive results. However, they use several hyperparameters that can be sensitive to changes in data or noise type~\cite{song2020learning, ortego2020multiobjective}. 

\textbf{Contrastive learning}: recent developments in self-supervised and contrastive learning~\cite{zhang2020decoupling,ortego2020multiobjective, li2020mopro} inspire new approaches in NLL.~\citet{li2020mopro} employed features learned by contrastive learning to detect out-of-distribution samples.

\subsection{Contrastive learning for vision data}
Contrastive learning extracts features by comparing each data sample with different samples. The central idea is to bring different instances of the same input image closer and spread instances from different images apart. The inputs are usually divided into positive (similar inputs) and negative pairs (dissimilar inputs). Frameworks have been recently developed, such as CPCv2~\cite{henaff2020data}, SimCLR~\cite{chen2020simple}, Moco~\cite{he2020momentum}. Once the self-supervised model is trained, the extracted representations can be used for downstream tasks.
In this work, the representations are used for noisy label classification.

\citet{chen2020simple} demonstrate that large sets of negatives (and large batches) are crucial in learning good representations.
However, large batches are limited by GPU memory. Maintaining a memory bank accumulating a large number of negative representations is an elegant solution decoupling the batch size from the number of negatives~\cite{misra2020self}. Nevertheless,
the representations get outdated in a few iterations. The Momentum Encoder~\cite{he2020momentum} addresses the issues by generating a dynamic memory queue of representations.
Other strategies aim at getting more meaningful negative samples to reduce the memory/batch size~\cite{kalantidis2020hard}.

%% file: sections/03_prelim.tex
\section{Preliminaries}
Let $D = \{ (\boldsymbol{x_i}, \overline{y_i}) \}_{i=1..n}, \boldsymbol{x_i} \in \mathbb{R}^{d}, \overline{y_i} \in \{1, \cdots , K\}$ denote a noisy input dataset with an unknown number of samples incorrectly labelled. The associated true and unobservable labels are written $y_i$. The images $\boldsymbol{x_i}$ are of size $d$ and the classification problem has $K$ classes. The goal is to train a deep neural network (DNN) $f$. Using a robust loss function for training consists of minimizing the empirical risk defined by robust loss functions in order to find the set of optimal parameters $ \theta $.
The one-hot encoding of the label is denoted by the distribution $q(k |\bm{x})$ for a sample $\bm{x}$ and a class $k$, such as $ q(y_i | \bm{x_i}) =1$ and $ q(k \ne y_i | \bm{x_i}) = 0, \: \forall i \in \{ 1, \cdots , n \}$. The probability vector of $f$ is defined by the softmax function $p(k|\bm{x}) = \frac{e^{z_k}}{\sum_{j=1}^K e^{z_j}}$ where $z_k$ denotes the logits output with respect to class $k$.

\subsection{Classification with robust loss functions}
The method employs noise-robust losses to train the classifier in the presence of noisy labels. Such losses improve the classification accuracy compared to the commonly used Cross Entropy (CE), as illustrated in~\autoref{fig:broad_loss_comparison}. In this section, the general empirical risk for a given mini-batch is defined by $ L = \sum_{i=1}^N \mathcal{L}(f(x_i),\overline{y_i}) =  \sum_{i=1}^N l_i $ . The term $l_i$ is modified by each loss function.

The classical CE is used as a baseline loss function not robust to noisy labels~\cite{ghosh2017robust} and is defined as:
\begin{equation}
    l_{ce} = - \sum_{k=1}^{K} q(k | \bm{x_i}) log(p(k | \bm{x_i})).
\end{equation}

As presented in~\autoref{sec:related_works},~\citet{ma2020normalized} introduce robust loss functions called Active Passive Losses that do not suffer from underfitting. We investigate the combination between the Normalized Focal Loss (NFL) and the Reversed Cross Entropy (RCE) called NFL+RCE. It shows promising results on various benchmarks. The NFL is defined as:
\begin{equation}
    l_{nfl} = \frac{ -\sum\limits_{k=1}^{K} q(k|\bm{x_i}) (1-p(k|\bm{x_i}))^\gamma log(p(k|\bm{x_i}))}{-\sum\limits_{j=1}^K \sum\limits_{k=1}^{K} q(y=j|\bm{x_i}) (1-p(k|\bm{x_i}))^\gamma log(p(k|\bm{x_i}))},
\end{equation}
where $\gamma \geq 0$ is an hyperparameter.
 The RCE loss is:
 \begin{equation}
     l_{rce} = -\sum_{k=1}^{K} p(k |\bm{x_i})log \left( q(k | \bm{x_i}) \right).
 \end{equation}
The final combination following the framework simply gives a different $\alpha$ and $\beta$ to each loss:
\begin{equation}
    l_{nfl+rce} = \alpha .l_{nfl} + \beta. l_{rce}.
\end{equation}
The two hyperparameters $\alpha$ and $\beta$ control the balancing between more active learning and less passive learning. For simplicity, $\alpha$ and $\beta$ are set to 1.0 without any tuning.

\citet{liu2020early} propose another framework to deal with noisy annotations based on the “early learning” phase. The loss, called Early Learning Regularization (ELR), adds a regularization term to capitalize on early learning. ELR is not strictly speaking a robust loss but belongs to robust penalization and label correction methods. The penalization term corrects the CE based on estimated soft labels identified with semi-supervised learning techniques. It prevents memorization of false labels by steering the model towards these targets. The regularization term maximizes the inner product between model outputs and targets:
\begin{equation}
 l_{elr} = l_{ce} + \frac{\lambda_{elr}}{N} log \left( 1 - \sum_{k=1}^{K} p(k|\bm{x_i}) t(k|\bm{x_i} )     \right). 
\end{equation}
The target is not set equal to the model output but is estimated with a temporal ensembling from semi-supervised methods. Let $t(k|\bm{x_i})^{(l)}$ denote the target for example $\bm{x_i}$ at iteration $l$ of training with a momentum $\beta$: 
\begin{equation}
    t(k|\bm{x_i})^{(l)} = \beta t(k|\bm{x_i})^{(l-1)} + (1 - \beta)p(k | \bm{x_i})^{(l)}.
\end{equation}

\subsection{Contrastive learning}
Contrastive learning methods learn representations by contrasting positive and negative examples. A typical framework is composed of several blocks~\cite{falcon2020framework}:
\begin{itemize}
   \item Data augmentation: Data augmentation is used to decouple the pretext tasks from the network architecture.~\citet{chen2020simple} study broadly the impact of data augmentation. We follow their suggestion combining random crop (and flip), color distortion, Gaussian blur, and gray-scaling.
   \item Encoding: The encoder extracts features (or representation) from augmented data samples. A classical choice for the encoder is the ResNet model~\cite{he2016deep} for image data. The final goal of the contrastive approach is to find correct weights for the encoder.
   \item Loss function: The loss function usually combines positive and negative pairs. The Noise Contrastive Estimation (NCE) and its variants are popular choices. The general formulation for such loss function is defined for the i-th pair as~\cite{wu2018unsupervised}:
\begin{equation}
    L_i = -log\frac{exp(\bm{z_i}^T \bm{z_{j(i)}}/ \tau)}{\sum_{a \in A(i)}exp(\bm{z_i}^T \bm{z_{a}}/ \tau)}, \; \text{with} \; i \in I,
    \label{eq:contrastive_loss}
\end{equation}
where $\bm{z}$ is a feature vector, $I$ is the set of indexes in the mini-batch, $i$ is the index of the anchor, $j(i)$ is the index of an augmented version of the anchor source image, $A(i)= I\setminus \{i \}$, and $\tau$ is a temperature controlling the dot product. The denominator includes one positive and $K$ negative pairs.
   \item Projection head: That step is not used in all frameworks. The projection head maps the representation to a lower-dimensional space and acts as an intermediate layer between the representation and the embedding pairs.~\citet{chen2020simple, chen2020improved} show that the projection head helps to improve the representation quality.
\end{itemize}

%% file: sections/04_representation_classifier.tex
\section{A framework coupling contrastive learning and noisy labels}

\begin{figure}[!ht]
\vskip 0.1in
\begin{center}
\centerline{\includegraphics[width=0.70\columnwidth]{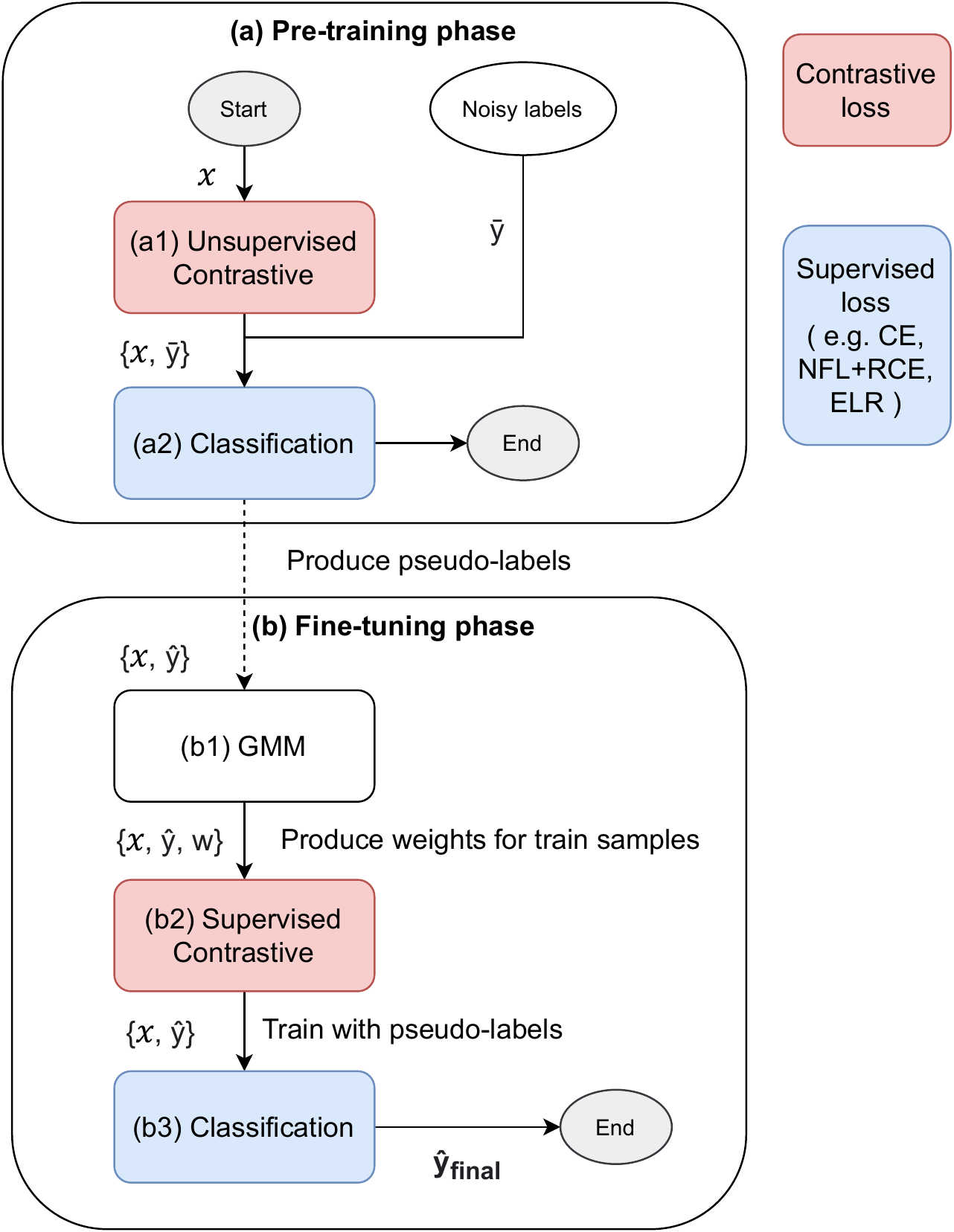}}
\caption{Overview of the framework consisting of two phases: pre-training (panel a) and fine-tuning (panel b). After a contrastive learning phase (a1) a classifier (a2) is trained to predict train-set pseudo-labels $\widehat{y}$. The fine-tuning phase uses $\widehat{y}$ as a new ground truth. First, a GMM model (b1) predicts the probability of correctness for each sample, used as a corrective weight factor in a supervised contrastive training (panel b2). The final predictions $\widehat{y}_{final}$ are produced by the (b3) classifier.}
\label{fig:full_workflow}
\end{center}
\vskip -0.15in
\end{figure}

As illustrated in~\autoref{fig:full_workflow}, our method classifies noisy samples in a two phased process. First, a classifier pre-trained with contrastive learning produces train-set pseudo-labels (pre-training phase, in panel a), used during the training of a subsequent fine-tuning phase (panel b). The underlying intuition is that the predicted pseudo-lables are more accurate than the original noisy labels. The contrastive learning performed in the first phase (panel a1) improves the performance the classifier (panel a2), sensitive to noisy labels; the resulting model can be also used in a standalone way with a reduced number of hyperparameters, without the underlying fine-tuning phase. 

The second phase leverages the pseudo-labels predicted by the pre-training in all underlying steps (b1-b3). To mitigate the effect of potentially incorrectly predicted pseudo-labels, a Gaussian Mixture Model (GMM, panel b1) with 2 components follows the small loss-trick to predict for each sample the probability of correctness. This value is used as a weight in a supervised contrastive step (panel b2), performed to improve the learned representations by taking advantage of the label information. A classification head is added to the contrastive model in order to produce the final predictions (panel b3). The fine-tuning phase can be seen as an adaptation of the pre-training phase to handle pseudo-labels.

To maximize the impact of the contrastive learning on the underlying classification, the supervised training is performed in 2 steps: a warm-up step, updating only the classifier layer (while keeping the encoder frozen) is followed by the full model training. We compared three different loss functions for the supervised classification: the classical CE, the robust NFL+RCE, and the ELR loss. 

\subsection{Sample selection and correction with pseudo-labels}
\begin{figure}[h]
\vskip 0.1in
\begin{center}
\centerline{\includegraphics[width=\columnwidth]{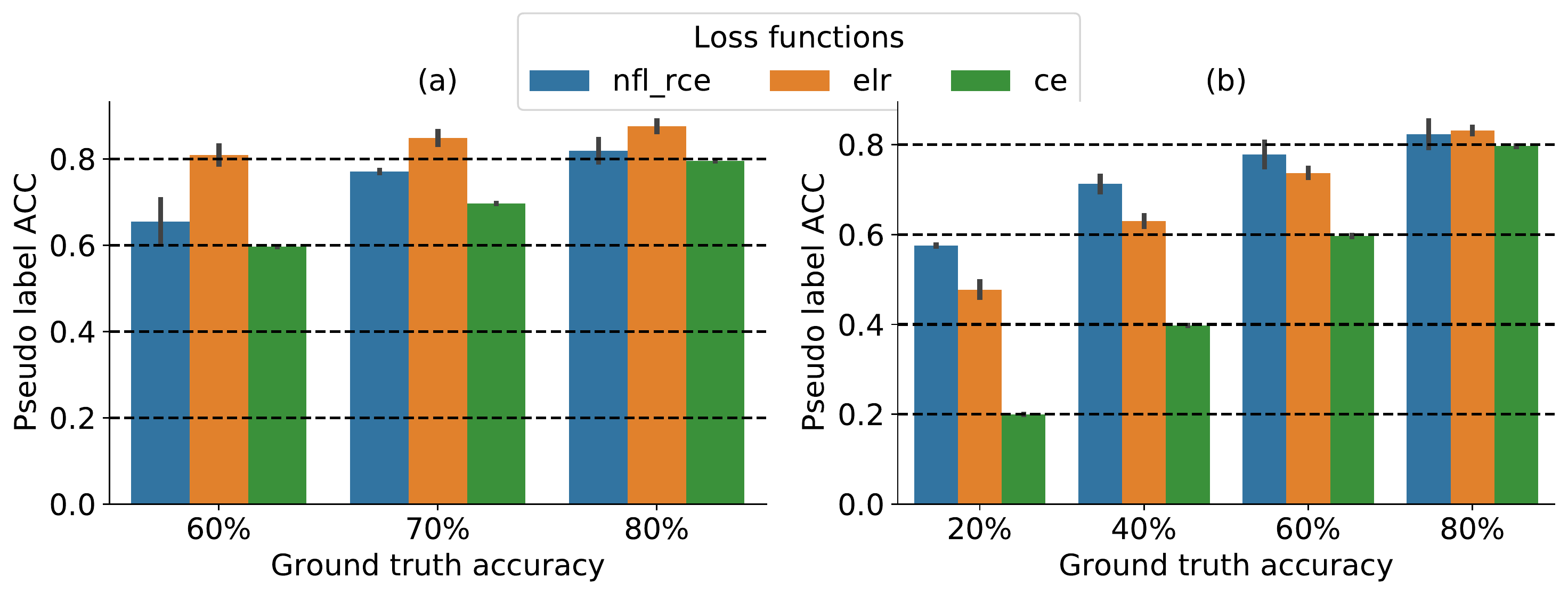}}
\caption{ Accuracy of pseudo labels on all simulated settings with asymmetric (a) and symmetric (b) noise, evaluated on CIFAR100. The correctness of the ground truth is represented on the x axis, while the accuracy of predicted pseudo labels on the y axis. In all experiments, the pseudo labels have a higher accuracy than the corrupted ground truth and this gain increases with the noise ratio}
\label{fig:pseudolabels}
\end{center}
\vskip -0.2in
\end{figure}

Pseudo labels represent one hot encoded model's predictions on the training set. Pseudo-labels were initially used in semi-supervised learning to produce annotations for unlabelled data; in the noisy label setting, various techniques (e.g. DivideMix, etc) identify a subset with a high likelihood of correctness and treat the remaining samples as the unlabeled counterpart in semi-supervised learning. In this work, we elaborate on the observation that the training set labels, predicted after training the model with a noise-robust loss function (i.e. the pseudo labels), are more accurate than the ground truth. This observation is supported by the results in~\autoref{fig:pseudolabels}, depicting the accuracy of pseudo labels predicted on CIFAR100, contaminated with various levels of asymmetric (panel a) and symmetric (panel b) noise. The pseudo labels are more accurate than the corrupted ground truth in both settings and bring a higher gain in performance as the noise ratio increases.

\begin{figure}[ht]
\vskip 0.1in
\begin{center}
\centerline{\includegraphics[width=0.7\columnwidth]{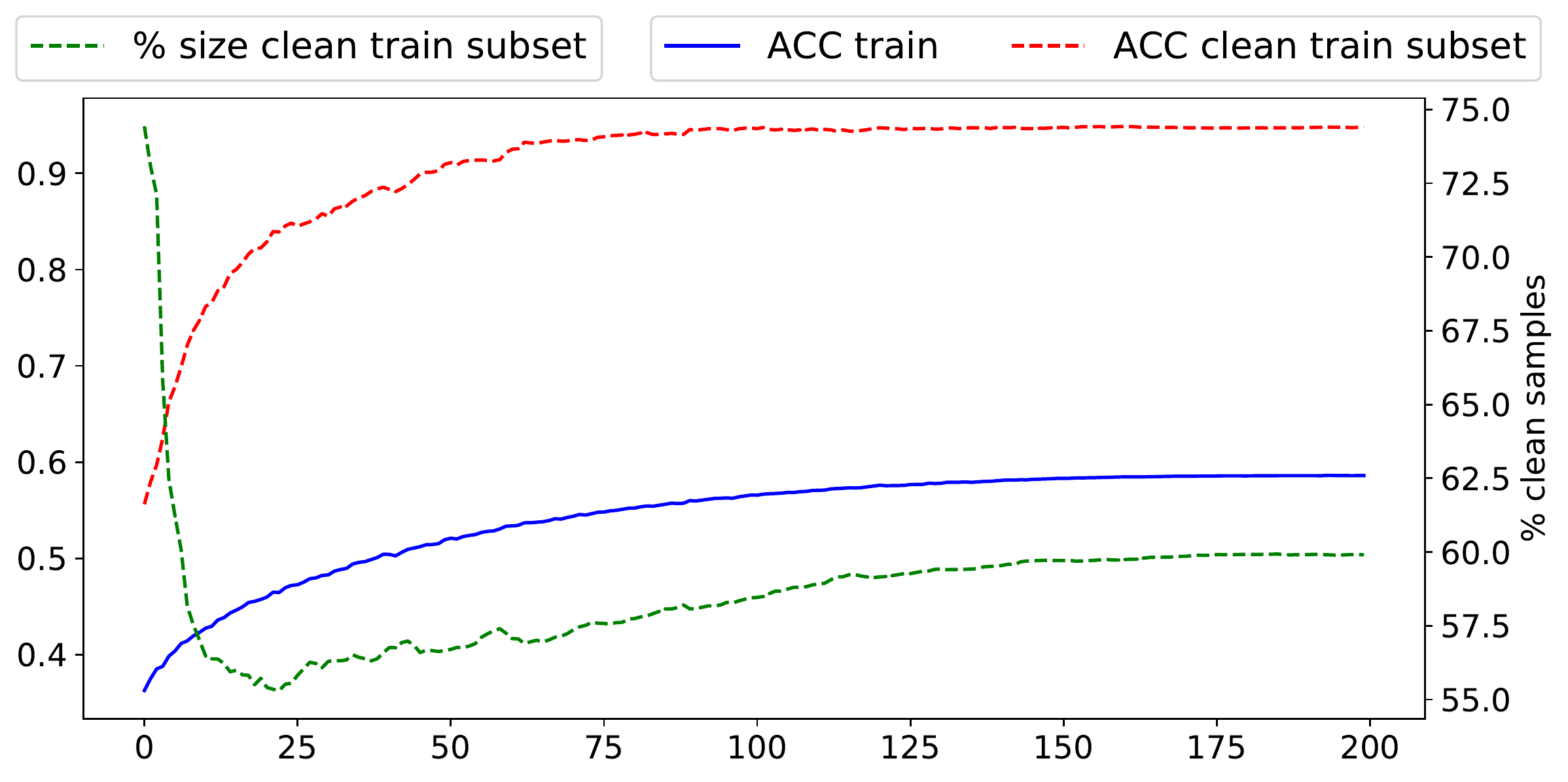}}
\caption{Accuracy of the entire training set (in blue) compared to the clean train subset (in red); The clean subset's percentual size is depicted in green. The example is performed on CIFAR100, with 40\% symmetric noise.} 
\label{fig:sample_weighting}
\end{center}
\vskip -0.2in
\end{figure}

As proposed by other approaches~\cite{arazo2019unsupervised}, the loss value on train samples can be used to discriminate between clean and mislabeled samples. The sample correctness probability is computed by fitting a 2 components GMM on the distribution of losses~\cite{li2020dividemix}. The underlying probability is used as a sample weight:
\begin{equation}
    w_i  = p(k= 0|l_i),
\end{equation}
where $l_i$ is the loss for sample $i$ and $k=0$ is the GMM component associated to the clean samples (lowest loss).
\autoref{fig:sample_weighting} depicts the evolution of the clean training set identified by GMM on an example: its accuracy grows from 0.6 to 0.93 while the size stabilizes at 60\% of the training set.

\subsection{Weighted supervised contrastive learning}
A modification to the contrastive loss defined in~\autoref{eq:contrastive_loss} has been proposed to leverage label information~\cite{khosla2020supervised}:
\begin{equation}
    L_i = -log\frac{1}{|P(i)|} \sum_{p \in P(i)}\frac{exp(\bm{z_i}^T \bm{z_{p}}/ \tau)}{\sum_{a\in A(i)}exp(\bm{z_i}^T \bm{z_a}/ \tau)},
    \label{eq:supervised_contrastive_loss}
\end{equation}
where $P(i) = \{ j \in I \setminus \{i\}, y_j = \widetilde{y_i} \}$ with $\widetilde{y_i}$ the prediction of the model for input $\bm{x_i}$. 

As explained in the previous section, the loss value for the training set samples is used to fit a GMM with 2 components, corresponding to correctly and incorrectly labeled samples. We adapted the supervised representation loss to employ $w$, a weighting factor representing the sample probability of membership to the correctly labeled component. Thus, likely mislabeled samples having large loss values would contribute only marginally to the supervised representations:
\begin{equation}
    L_i = -log\frac{1}{|P(i)|} \sum_{p \in P(i)}\frac{\widetilde{w_{p,i}} exp(\bm{z_i}^T \bm{z_{p}}/ \tau)}{\sum_{a \in A(i)}exp(\bm{z_i}^T \bm{z_p}/ \tau)},
    \label{eq:supervised_contrastive_loss_weighted}
\end{equation}
where $\widetilde{w_{p,i}}$ is a modified version of $w_p$ such as $\widetilde{w_{p,i}} = 1$ if $p = j(i)$ else $\widetilde{w_{p,i}}= w_i$. If all samples are considered as noisy,~\autoref{eq:supervised_contrastive_loss_weighted} is simplified into the classical unsupervised contrastive loss in~\autoref{eq:contrastive_loss}.

%% file: sections/06_experiments.tex
\section{Experiments}
The framework is assessed on three benchmarks and the contribution of each block identified in~\autoref{fig:full_workflow} is analyzed.

\subsection{Datasets}
\textbf{CIFAR10} and \textbf{CIFAR100}~\cite{krizhevsky2009learning}. These experiments assess the accuracy of the method against synthetic label noise. The two datasets are contaminated with simulated symmetric or asymmetric label noise reproducing the heuristic in~\citet{ma2020normalized}. The symmetric noise consists in corrupting an equal arbitrary ratio of labels for each class. The noise level varies from $0.2$ to $0.8$. For asymmetric noise~\cite{patrini2017making, liu2020early}, sample labels have been flipped within a specific set of classes, thus providing confusion between predetermined pairs of labels.
For CIFAR100, 20 groups of super-classes have been created, each consisting of 5 sub-classes. The label flipping is performed only within each super-class circularly. The asymmetric noise ratio is explored between $0.2$ and $0.4$.

\textbf{Webvision}~\cite{li2017webvision}. This is a real-world dataset with noisy labels. It contains 2.4 million images crawled from the web (Google and Flickr) that share the same 1,000 classes from the ImageNet dataset. The noise ratio varies from 0.5\% to 88\%, depending on the class. In order to speed-up the training time, we used mini Webvision~\cite{jiang2018mentornet}, consisting of only top 50 classes in the Google subset (66,000 images).

\textbf{Clothing1M}~\cite{xiao2015learning}. Clothing 1M is a large real-world dataset consisting of 1 million images on 14 classes of clothing articles. Being gathered from e-commerce websites, Clothing1M embeds an unknown ratio of label noise. Additional validation and test sets, consisting of 14k and 10k clean labeled samples have been made available. In order to speed-up the training time, we selected a subset of 56,000 images keeping the initial class distribution.

Both Webvision and Clothing1M images were resized to $128 \times 128$. Therefore, the reported results may differ from other papers cropping the images to a $224\times 224$ resolution.

\subsection{Settings}
We use the contrastive SimCLR framework~\cite{chen2020simple} with a ResNet18~\cite{he2016deep} (without ImageNet pre-training) as encoder. A projection head was added after the encoder for the contrastive learning with the following architecture: a multi-layer perceptron with one hidden layer and a ReLu non-linearity. The classifier following the contrastive learning step has a simple multilayer architecture: a single hidden layer with batch normalization and a ReLU activation function. A comparison with a linear classifier is provided in the supplementary materials.

For all supervised classification, we use SGD optimizer with momentum 0.9 and cosine learning rate annealing. The NFL hyperparameter $\gamma$ is set to $0.5$. Unlike the original paper, the ELR hyperparameters do no depend on the noise type: the regularization coefficient $\lambda_{elr}$ and the momentum $\beta$ are set to $3.0$ and $0.7$. Details on the experiment setting can be found in the supplementary materials.

All codes are implemented in the PyTorch framework~\cite{pytorch2019}. The experiments for CIFAR are performed with a single Nvidia TITAN V-12GB and the experiments for Webvision and Clothing1M are performed with a single Nvidia Tesla V100-32GB, demonstrating the accessibility of the method. Our implementation has been made available along with the supplementary materials.

\section{Results}
All experiments presented in this secion evaluate our method's performance with the top-1 accuracy score. 
\subsection{Impact of contrastive pre-training}
To evaluate the impact of the contrastive pre-training on the classification model, the proposed method (pre-training phase) is compared with a baseline classifier, trained for 200 epochs without contrastive learning. For each simulated dataset, we compare robust losses (e.g. NLF+RCE and ELR) and cross entropy. Results for CIFAR10 and CIFAR100 are depicted in~\autoref{tab:results_block1_cifar} for different levels of symmetric and asymmetric noise. The pre-training improves the accuracy of the three different baselines for both datasets with different types and ratios of label noise. The largest differences are observed for the noisiest case with $80\%$ noise. The pre-training outperforms the baselines by large margins between $10$ and $75$ for CIFAR10 and between $5$ and $30$ for CIFAR100.

\begin{table}[ht]
\caption{Results on both CIFAR10 and CIFAR100 using symmetric noise (0.2 - 0.8) and asymmetric noise (0.2 - 0.4). We compare training from scratch or from pre-trained representation. Best scores are in bold for each noise scenario and each loss.}
\label{tab:results_block1_cifar}
\vskip 0.15in
\begin{center}
\begin{tabular}{p{6.0mm}p{4mm}p{7.0mm}p{8.5mm}p{8.5mm}p{8.5mm}p{8.5mm}}
\cline{4-7}
                      &                         &          & \multicolumn{2}{c}{CIFAR10} & \multicolumn{2}{c}{CIFAR100}  \\ \hline
Type                  & $\eta$ & Loss     & Base    & Pre-t.          & Base         & Pre-t.       \\ \hline
\multirow{12}{*}{Sym} & \multirow{3}{*}{0.2}    & ce       & 77.2     & \textbf{87.7}    & 55.6          & \textbf{56.5} \\ \cline{3-7} 
                      &                         & elr      & 90.3     & \textbf{93.0}    & 64.1          & \textbf{67.4} \\ \cline{3-7} 
                      &                         & nfl+rce & 91.0     & \textbf{92.7}    & 66.6          & \textbf{68.8} \\ \cline{2-7} 
                      & \multirow{3}{*}{0.4}    & ce       & 58.2     & \textbf{78.0}    & 39.9          & \textbf{41.9} \\ \cline{3-7} 
                      &                         & elr      & 82.3     & \textbf{92.0}    & 56.9          & \textbf{62.0} \\ \cline{3-7} 
                      &                         & nfl+rce & 87.0     & \textbf{91.4}    & 60.2          & \textbf{66.3} \\ \cline{2-7} 
                      & \multirow{3}{*}{0.6}    & ce       & 35.2     & \textbf{59.2}    & 21.8          & \textbf{26.8} \\ \cline{3-7} 
                      &                         & elr      & 64.2     & \textbf{90.4}    & 40.6          & \textbf{55.7} \\ \cline{3-7} 
                      &                         & nfl+rce & 80.2     & \textbf{88.1}    & 47.0          & \textbf{61.8} \\ \cline{2-7} 
                      & \multirow{3}{*}{0.8}    & ce       & 17.0     & \textbf{27.3}    & 7.80          & \textbf{12.4} \\ \cline{3-7} 
                      &                         & elr      & 18.3     & \textbf{84.8}    & 16.2          & \textbf{45.3} \\ \cline{3-7} 
                      &                         & nfl+rce & 42.8     & \textbf{59.9}    & 20.1          & \textbf{50.2} \\ \hline
\multirow{9}{*}{Asym} & \multirow{3}{*}{0.2}    & ce       & 84.0     & \textbf{87.9}    & \textbf{59.0} & 57.8          \\ \cline{3-7} 
                      &                         & elr      & 91.8     & \textbf{92.4}    & \textbf{70.3} & 70.2          \\ \cline{3-7} 
                      &                         & nfl+rce & 90.2     & \textbf{91.5}    & 63.9          & \textbf{68.4} \\ \cline{2-7} 
                      & \multirow{3}{*}{0.3}    & ce       & 79.2     & \textbf{83.9}    & \textbf{50.6} & 50.4          \\ \cline{3-7} 
                      &                         & elr      & 89.6     & \textbf{91.7}    & \textbf{69.8} & 69.3          \\ \cline{3-7} 
                      &                         & nfl+rce & 86.7     & \textbf{89.9}    & 53.5          & \textbf{63.5} \\ \cline{2-7} 
                      & \multirow{3}{*}{0.4}    & ce       & 75.3     & \textbf{77.8}    & 41.8          & \textbf{42.4} \\ \cline{3-7} 
                      &                         & elr      & 72.3     & \textbf{89.5}    & \textbf{67.6} & \textbf{67.6} \\ \cline{3-7} 
                      &                         & nfl+rce & 80.0     & \textbf{82.4}    & 40.6          & \textbf{47.8} \\ \hline
\end{tabular}
\end{center}
\vskip -0.15in
\end{table}

In addition to the comparisons with ELR and NFL+RCE, performed using our implementations (column Base in ~\autoref{tab:results_block1_cifar}), we present the results reported by other recent competing methods. As shown in the introduction, numerous contributions have been made to the field in the last years. Six recent representative methods are selected for comparison: Taks~\cite{song2020no}, Co-teaching+~\cite{yu2019does}, ELR~\cite{liu2020early}, DivideMix~\cite{li2020dividemix}, SELF~\cite{nguyen2019self}, and JoCoR~\cite{wei2020combating}. The results are presented in~\autoref{tab:acc_comparaison}. The difference between the scores reported by ELR and those obtained with our run (using the same implementation, but slightly different hyper-parameters and a ResNet18 instead of a ResNet34) suggests that the method is less stable on data contaminated with asymmetric noise and sensitive to small changes hyperparameters. Moreover, ELR proposes hyperparameters having different values depending on the type of dataset (i.e. CIFAR10/CIFAR100) and underlying noise (i.e. symmetric/asymmetric), identified after a hyperparameter search exercise. The best scores are reported by DivideMix and they surpass all other techniques. One can note DivideMix uses a PreAct ResNet18 while we use a classical ResNet18. Moreover, a recent study~\cite{ortego2020multiobjective} attempted to replicate these values and reported significantly lower results on CIFAR100 (i.e. $49.5\%$ instead of $59.6\%$ on symmetric data and $50.9\%$ instead of $72.1\%$ on asymmetric data). Our framework compares favourably with the other competing methods, both on symmetric and asymmetric noise.

\begin{table}[]
\caption{Accuracy scores compared with 6 methods (Taks, Co-teaching+, ELR, DivideMix, SELF, and JoCoR) on CIFAR10 (C10) and CIFAR100 (C100). The cases most affected by dropout are presented, with symmetric (S) and asymmetric (A) noise. Top-2 scores are in bold}
\label{tab:acc_comparaison}
\vskip 0.15in
\begin{center}
\begin{tabular}{lllll}
\cline{2-5}
                                                           & \begin{tabular}[c]{@{}l@{}}C10\\ 80\% S\end{tabular} & \begin{tabular}[c]{@{}l@{}}C10\\ 40\% A\end{tabular} & \begin{tabular}[c]{@{}l@{}}C100\\ 80\% S\end{tabular} & \begin{tabular}[c]{@{}l@{}}C100\\ 40\% A\end{tabular} \\ \hline
\begin{tabular}[c]{@{}l@{}}Ours \\ (ELR)\\ \end{tabular}  & \textbf{84.8}                                                & 89.5                                                 & \textbf{45.3}                                                  & 67.6                                                  \\ \hline
\begin{tabular}[c]{@{}l@{}}ELR~\cite{liu2020early}\end{tabular}       & 73.9                                                 & \textbf{91.1}                                                 & 29.7                                                  & \textbf{73.2}                                                  \\ \hline
\begin{tabular}[c]{@{}l@{}}Taks~\cite{song2020no}\end{tabular}      & 40.2                                                 & 73.4                                                 & 16.0                                                  & 35.2                                                  \\ \hline
\begin{tabular}[c]{@{}l@{}}Co-teach+~\cite{yu2019does}\end{tabular} & 23.5                                                 & 68.5                                                 & 14.0                                                  & 34.3                                                  \\ \hline
\begin{tabular}[c]{@{}l@{}}DivideMix~\cite{li2020dividemix}\end{tabular} & \textbf{92.9}                                                 & \textbf{93.4 }                                                & \textbf{59.6}                                                  & \textbf{72.1}                                                  \\ \hline
\begin{tabular}[c]{@{}l@{}}SELF~\cite{nguyen2019self}\end{tabular}      & 69.9                                                 & 89.1                                                 & 42.1                                                  & 53.8                                                  \\ \hline
\begin{tabular}[c]{@{}l@{}}JoCoR~\cite{wei2020combating}\end{tabular}     & 25.5                                                 & 76.1                                                 & 12.9                                                  & 32.3                                                 
\end{tabular}
\end{center}
\end{table}

Webvision and Clothing1M results are presented in~\autoref{tab:results_block1_real}. The contrastive framework outperforms the respective baselines for the three loss functions. Because the images have a reduced size, and for Clothing1M, we use a smaller training set, the direct comparison with competing methods is less relevant. However, the observed gap in performance is significant and promising for training images with higher resolution. Moreover, a ResNet50 model has been trained with our framework on the Webvision dataset with a higher resolution ($224\times 224$). The accuracy reaches respectively $75.7\%$ and $76.2\%$ for CE and ELR. These results are very close to the values reported with DivideMix ($77.3\%$) and ELR+ ($77.8\%$) using a larger model, Inception-ResNet-v2 (the difference is more than $4\%$ on the ImageNet benchmark~\cite{bianco2018benchmark}).
\begin{table}[ht]
\caption{Top-1 accuracy for mini-Webvision and Clothing1M. Best scores are in bold for each dataset and each loss. Pre-t represents the pre-training phase while Fine-tune refers to the results after the fine-tuning step.}
\label{tab:results_block1_real}
\vskip 0.15in
\begin{center}
\begin{tabular}{p{7.5mm}p{5.0mm}p{7.8mm}p{7.0mm}p{5.0mm}p{7.8mm}p{7.0mm}}
\cline{2-7}
         & \multicolumn{3}{c}{Webvision}&\multicolumn{3}{c}{Clothing1M} \\ \hline
Loss     & Base.   & Pre-t. & Fine-tune & Base.   & Pre-t. & Fine-tune\\ \hline
ce       & 51.8      & 57.1& \textbf{58.4}&  54.8    & 59.1&  \textbf{61.5} \\ \hline
elr      & 53.0      & 58.1& \textbf{59.0}& 57.4     & \textbf{60.8}  &  60.4\\ \hline
nfl+rce & 49.9     & 54.8& \textbf{58.2}&   57.4   &  59.4 & \textbf{60.1}\\ \hline
\end{tabular}
\end{center}
\vskip -0.15in
\end{table}

Supported by this first set of experiments, the preliminary pre-training with contrastive learning shows great performances. The accuracy of both traditional and robust-loss classification models is significantly improved.

\subsection{Sensitivity to the hyperparameters}
\begin{figure}[ht]
\vskip 0.1in
\begin{center}
\centerline{\includegraphics[width=0.7\columnwidth]{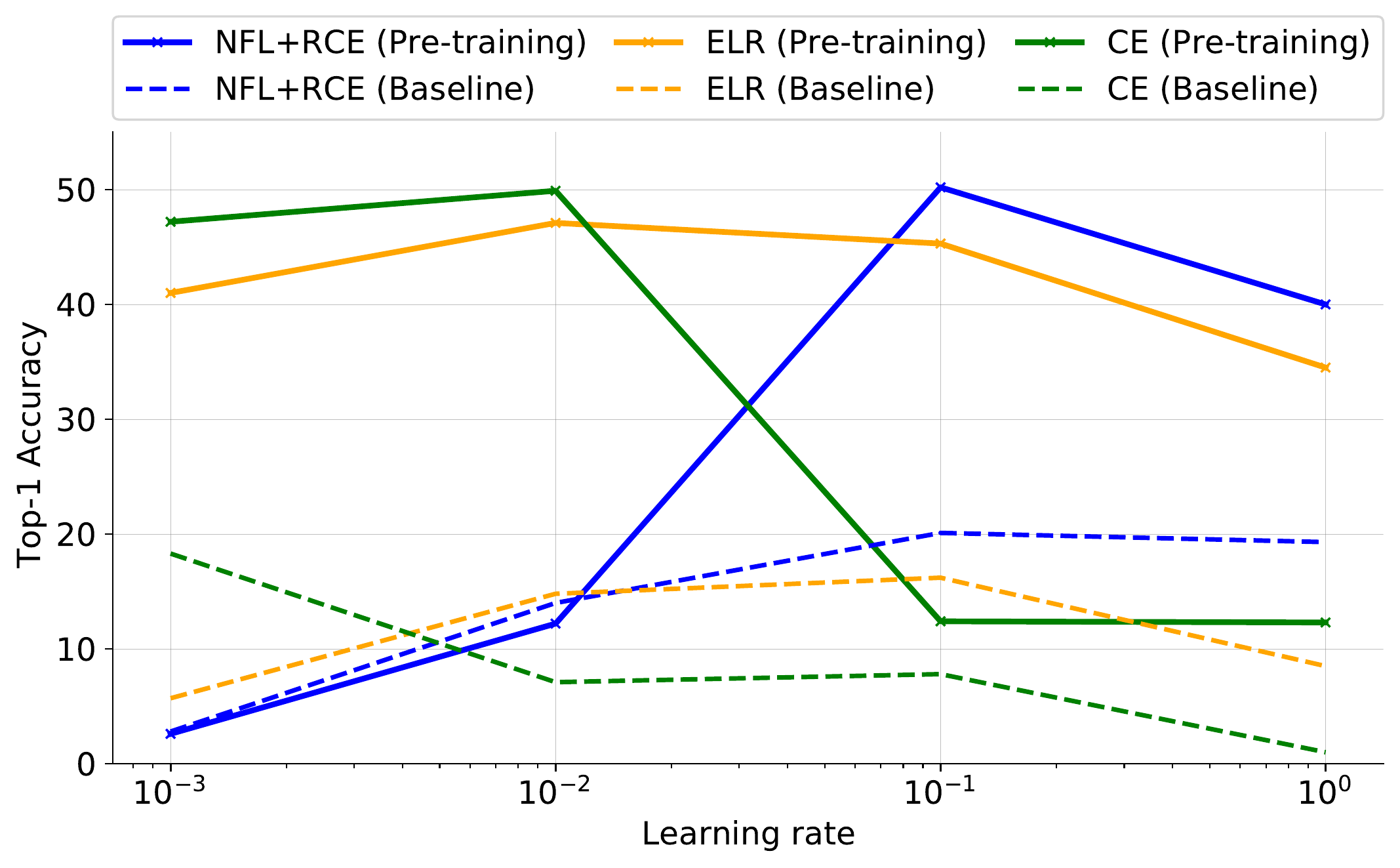}}
\caption{Learning rate sensitivity for CIFAR100 with 80\% noise. The explored learning rate values are $\{0.001, 0.01, 0.1, 1.0\}$. The baseline (dashed line) is compared with our framework (solid line).}
\label{fig:sensitivity}
\end{center}
\vskip -0.2in
\end{figure}

Estimating the best hyperparameters is complex for datasets with noisy labels as clean validation sets are not available. For instance,~\citet{ortego2020multiobjective} show that two efficient methods (eg. ELR and DivideMix) could be sensitive to specific hyperparameters. Therefore a hyperparameter sensitivity study has been carried out to estimate the stability of the framework for the learning rate.~\autoref{fig:sensitivity} depicts the sensitivity on CIFAR100 with $80\%$ noise. CE and NFL+RCE seem to have opposite behaviors. The CE reaches competitive results with small learning rates but is prompt to overfitting for higher learning rates. The NFL+RCE loss tends to underfitting for the lowest learning rates but is quite robust for higher values. The ELR loss has the smallest sensitivity to the learning for the investigated range but does not reach the best values obtained with CE or NFL+RCE. We can assume that the regularization term coupled with pre-training is very efficient. It prevents memorization of the false labels as observed with CE. Results for other noise ratios have been documented in the supplementary materials.

This sensitivity analysis is limited to the learning rate. Investigating the impact of other hyperparameters, such as the momentum $\beta$ or the regularization factor $\lambda_{elr}$, could be interesting. In their original papers, ELR and NFL+RCE reach respectively $25.2\%$ and $30.3\%$ with other hyperparameters. These values are still far from the improvements brought by the contrastive pre-training but it suggests that the results could be improved with different hyperparameters.

Our empirical results indicate that the analyzed methods may be sensitive to hyperparameters. Despite the promised robustness to label noise, the analyzed robust losses are also affected by overfitting or underfitting. Our experiments have been built upon the parameters recommended in each issuing paper (e.g. ELR, SIMCLR) but, since the individual building blocks can be affected by small variations in input parameters, the performance of our method may also be impacted. Finding a relevant method to estimate proper hyperparameters in NLL remains a challenge. In the absence of a clean validation set, identifying when overfitting starts also remains an open challenge. This is demonstrated by our studies on the behaviour of the (also noise-corrupted) validation set and another two recently proposed methods, analyzing the stability of the loss function on the train set and the changes in the upstream layers. These experiments are detailed in Supplementary Materials.

\subsection{Impact of the fine-tuning phase}
\begin{figure}[ht]
\vskip 0.2in
\begin{center}
\centerline{\includegraphics[width=\columnwidth]{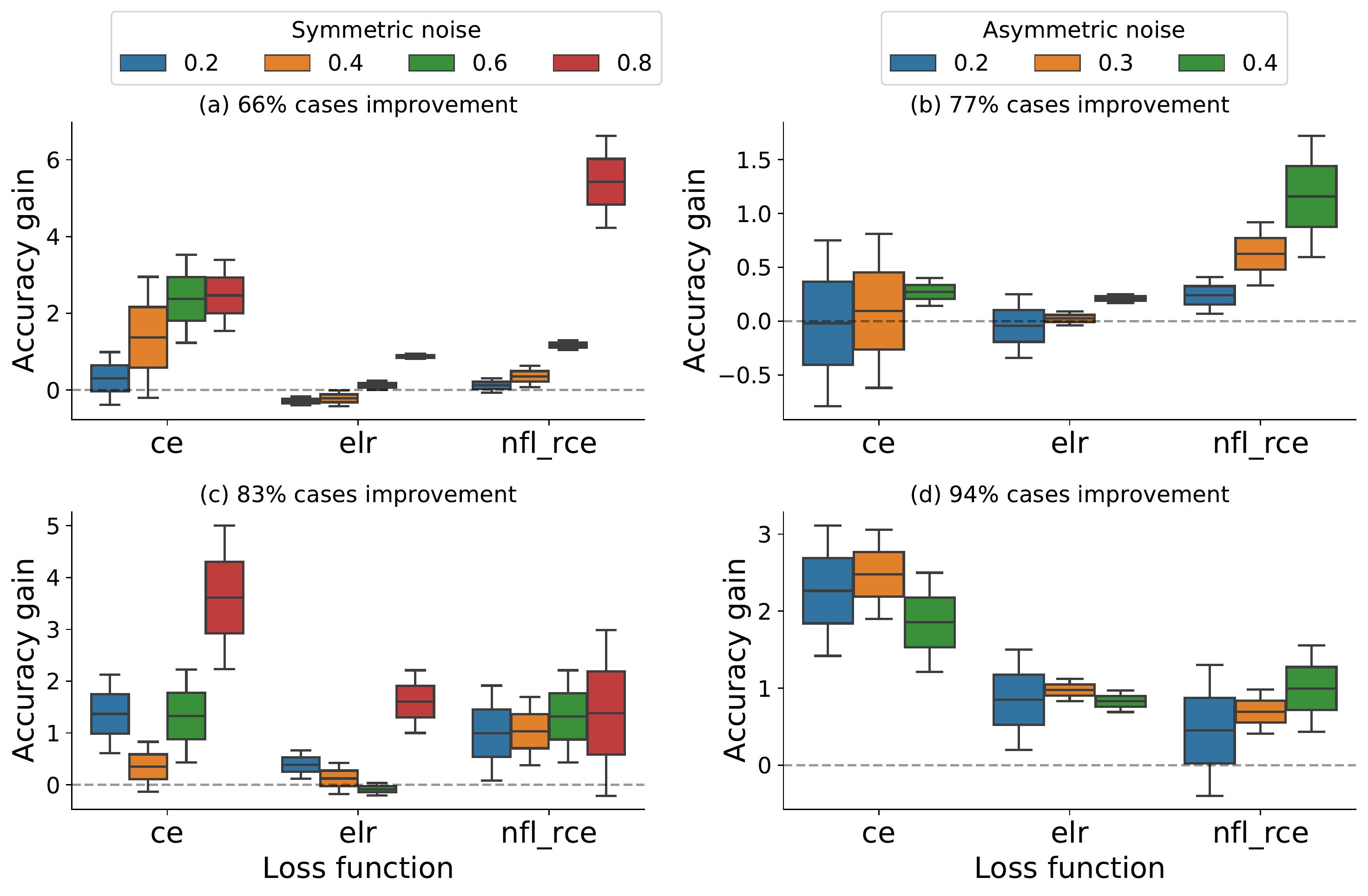}}
\caption{Accuracy gain when performing the fine-tuning phase after the pre-training block (computed as the difference between fine-tuning accuracy and pre-training accuracy). The plot gathers the results for all noise ratios on CIFAR10 (panels a, b) and CIFAR100 (c, d) with symmetric (first column) and asymmetric (second column) noise.} 
\label{fig:second_classification}
\end{center}
\vskip -0.2in
\end{figure}

Experimental results on synthetic label noise, depicted in~\autoref{fig:second_classification}, show that continuing the presented pre-training block (\autoref{fig:full_workflow}) with the fine-tuning phase increases the accuracy in over 65\% of cases on CIFAR10 and over 80\% of cases on CIFAR100. For both datasets, asymmetric noise data benefit more from this approach than symmetric noise. All experiments only use the input parameters proposed in the loss-issuing papers.

The sample selection has also got a positive impact on the two real-world datasets, as shown in~\autoref{tab:results_block1_real} by the "Fine-tune" columns. The average accuracy improvement is about $1.8\%$. Only the ELR loss function slightly decreases the performance on Clothing1M. 

 Enriching pretrained models with sample weighting and selection, pseudo labels instead of corrupted targets, and supervised contrastive pre-training can improve the classification accuracy. However, such an approach raises the question of a trade-off between complexity, accuracy improvement, and computation time.


\section{Discussion and limits of the framework}
\begin{figure}[ht]
\vskip 0.1in
\begin{center}
\centerline{\includegraphics[width=\columnwidth]{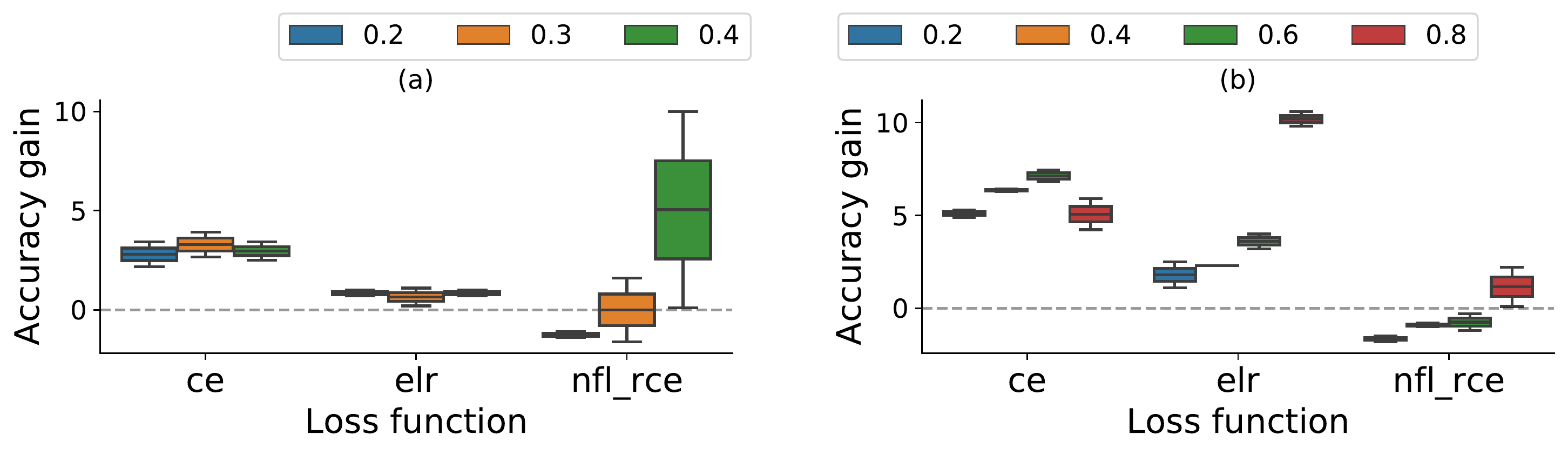}}
\caption{Top-1 accuracy gain for the dynamic bootstrapping on CIFAR100 with asymmetric (a) and symmetric noise (b). Dynamic bootstrapping is an alternative to the proposed fine-tuning phase.} 
\label{fig:mixup}
\end{center}
\vskip -0.2in
\end{figure}

In addition to the presented fine-tuning phase, we evaluated the performance of other promising techniques, such as the dynamic bootstrapping with mixup~\cite{arazo2019unsupervised}. This strategy has been developed to help convergence under extreme label noise conditions. Details can be found in the supplementary materials. The improvement that dynamic bootstraping can bring when used after pre-training is depicted in~\autoref{fig:mixup}. In most of the cases, this technique improves the accuracy, as indicated by the positive accuracy gain scores, measuring the difference between the accuracy after dynamic bootstraping and the accuracy of the pre-training phase. ELR and CE benefit most from this addition for CIFAR100. The impact of the dynamic boostrapping should also be analyzed for the fine-tuning phase and for larger datasets, such as Webvision or Clothing1M.



One of the major drawbacks of our method is the extra computational time needed to learn representations with contrastive learning. A detailed study, comparing the execution time of our framework with 6 other competing methods has been provided in supplementary materials. The pre-training phase doubles the execution time of a reference baseline, consisting of performing only a single classification step, while the entire framework increases the execution time 3 to 4 times the baseline value. However, the constrastive learning does not increase the need for GPU memory if the batch size is limited for the contrastive learning~\cite{mitrovic2020less, he2020momentum}. The computational time could be reduced by initializing the contrastive step with the pretrained weights from ImageNet.

Most state-of-the-art approaches also leverage computationally expensive settings, consisting of larger models (e.g. ResNet50), dual model training, or data augmentation such as mixup. In this work, we explored the limits of a restricted computational setting, consisting of a single GPU and 8GB RAM. All experiments use a ResNet18 model, batch sizes of $256$, and for real-world datasets, the images have been rescaled (e.g. $128\times128$ instead of $224\times224$). We also foresee that the constrastive learning step could be improved by images with higher resolutions as smaller details could be identified in the representation embedding.

There remain multiple open problems for future research, such as: i) identifying the start of the memorization phase in the absence of a clean dataset, ii) studying the impact of contrastive learning on other models for noisy labels such as DivideMix, iii) comparing SimCLR approach in the context of noisy labels with other contrastive frameworks (the impact of Moco is studied in the supplementary materials) and other self-supervised approaches, and iv) having a better theoretical understanding of the interaction between the initial state precomputed with contrastive learning and the classifier in presence of noisy labels. Moreover, the analysis carried out in this work should be validated on larger settings, in particular on Clothing1M with a ResNet50, higher resolutions, and the full dataset.

%% file: sections/07_conclusion.tex
\section{Conclusions}
In this work, we presented a contrastive learning framework optimized with several adaptations for noisy label classification. Supported by an extensive range of experiments, we conclude that a preliminary representation pre-training improves the performance of both traditional and robust-loss classification models. Additionally, multiple techniques can be used to fine-tune and further optimize these results; however, no approach provides a significant improvement systematically on all types of datasets and label noise. The cross-entropy penalized by Early-Learning Regularization (ELR) shows the best overall results for synthetic noise but also real-world datasets. 

However, the training phases remain sensitive to input configuration. Overfitting is the common weakness of all studied models. When trained with tuned parameters, even traditional (cross-entropy) models provide competitive results, while robust-losses are less sensitive. The typical noisy label adaptations, such as sample selection or weighting, the usage of pseudo labels, or supervised contrastive losses, improve the performance to a lesser extent but increase the framework's complexity. We hope that this work will promote the use of contrastive learning to improve the robustness of the classification process with noisy labels.